 \documentclass[5p,times]{elsarticle}
\pdfoutput=1

\usepackage{amsmath,amssymb,amsfonts}
\usepackage{algorithmic}
\usepackage{graphicx}
\usepackage{textcomp}
\usepackage[ruled,vlined]{algorithm2e}
\usepackage{multirow}
\usepackage{xspace}
\usepackage{subfigure}
\usepackage{graphicx} 
\usepackage{multirow}
\usepackage[table,xcdraw]{xcolor}
 \usepackage{enumitem}
\usepackage{tablefootnote}
\usepackage{url}

\newcommand\zbabb{{Our}\xspace}

\definecolor{crimson}{rgb}{0, 0, 0}

\journal{Future Generation Computer Systems}

\begin{document}

\begin{frontmatter}



\title{Latency-Constrained DNN Architecture Learning for Edge Systems using Zerorized Batch Normalization}

\author[label1,label2]{Shuo Huai}
\ead{shuo001@e.ntu.edu.sg}
\author[label4]{Di Liu}
\ead{liu.di@ntu.edu.sg}
\author[label1,label2]{Hao Kong}
\ead{hao004@e.ntu.edu.sg}
\author[label1]{Weichen Liu\corref{cor1}}
\ead{liu@ntu.edu.sg}
\author[label3]{Ravi Subramaniam}
\ead{ravi.subramaniam@hp.com}
\author[label3]{Christian Makaya}
\ead{christian.makaya@hp.com}
\author[label3]{Qian Lin}
\ead{qian.lin@hp.com}
\affiliation[label1]{organization={School of Computer Science and Engineering},
            addressline={Nanyang Technological University},
            postcode={639798},
            country={Singapore}}
\affiliation[label2]{organization={HP-NTU Digital Manufacturing Corporate Lab},
            addressline={Nanyang Technological University},
            postcode={639798},
            country={Singapore}}
\affiliation[label4]{organization={Department of Computer Science},
  addressline={Norwegian University of Science and Technology},
            postcode={7491},
            city={Trondheim},
            country={Norway}}
\affiliation[label3]{organization={HP Inc.},
            city={Palo Alto},
            postcode={94304},
            state={California},
            country={United States}}

\cortext[cor1]{Corresponding author.}
\begin{abstract}
Deep learning applications have been widely adopted on edge devices, to mitigate the privacy and latency issues of accessing cloud servers. Deciding the number of neurons during the design of a deep neural network to maximize performance is not intuitive. Particularly, many application scenarios are real-time and have a strict latency constraint, while conventional neural network optimization methods do not directly change the temporal cost of model inference for latency-critical edge systems. In this work, we propose a latency-oriented neural network learning method to optimize models for high accuracy while fulfilling the latency constraint. For efficiency, we also introduce a universal hardware-customized latency predictor to optimize this procedure to learn a model that satisfies the latency constraint by only a one-shot training process. The experiment results reveal that, compared to state-of-the-art methods, our approach can well-fit the `hard' latency constraint and achieve high accuracy. Under the same training settings as the original model and satisfying a 34ms latency constraint on the ImageNet-100 dataset, we reduce GoogLeNet's latency from 40.32ms to 34ms with a 0.14\% accuracy reduction on the NVIDIA Jetson Nano. When coupled with quantization, our method can be further improved to only 0.04\% drop for GoogLeNet. On the NVIDIA Jetson TX2, we compress VGG-19 from 119.98ms to 34ms and even improve its accuracy by 0.5\%, and we scale GoogLeNet up from 20.27ms to 34ms and achieve higher accuracy by 0.78\%. We also open source this framework at \textcolor{blue}{\url{https://github.com/ntuliuteam/ZeroBN}}.

\end{abstract}



\begin{keyword}


deep neural network \sep latency optimization \sep edge device \sep latency prediction \sep  neural network learning \sep high performance

\end{keyword}

\end{frontmatter}


\section{Introduction}\label{section:introduction}
Deep Neural Networks (DNNs) have achieved striking success in a variety of applications, including image recognition and object detection \cite{NetworkSlimming}. Meanwhile, as accessing cloud servers hurts data confidentiality and cannot guarantee latency owing to unstable network bandwidth, many DNN applications have been increasingly deployed on edge devices such as autonomous vehicles, healthcare devices and so on \cite{DLSurvey}. Although various model architectures are designed by manual \cite{googlenet} or by neural architecture search (NAS) \cite{nas} to increase the accuracy and improve the efficiency of DNN applications, the emergence of diverse edge devices with different computational capabilities and resources results in a mismatch between existing models and the majority of edge devices.


Many applications have rigorous response latency requirements, such as self-driving vehicles and drone tracking. But complicated models significantly degrade the performance of edge devices, leading to them being unacceptable for latency-critical systems. On the other hand, simple models sacrifice too much accuracy to reduce the latency. Thus, to implement an efficient edge DNN system with the best accuracy and guaranteed latency, DNN models need to contain efficient architectures and appropriate computational costs (floating-point operations per second, FLOPs) to fully utilize the computational resource provided by the underlying hardware.

Although some works use NAS \cite{nas}  to solve this problem, designing new DNN models for each edge device is very time-consuming and requires significant engineering efforts \cite{nas, wan2020fbnetv2}. Therefore, optimizing existing DNN models is more promising and versatile \cite{SoftPrune}. \textcolor{black}{Neural network pruning \cite{NetworkSlimming, SoftPrune,FPGM, PGMPF, oto} and Neural network scaling \cite{tan2019efficientnet, kong2022hacscale} } can change the model complexity by removing redundant parameters from DNNs or extending DNNs with more channels and layers, respectively. Thus, we borrow the idea from them to improve the model architecture and change the inference latency while keeping high accuracy.



Nonetheless, these existing methods suffer from \textbf{three} flaws. Firstly, these methods adopt indirect performance metrics, for example,  FLOPs or the number of channels and layers as the optimization target. These metrics are not hardware-specific and they cannot accurately reflect the real latency of a DNN model on edge devices. These present technologies are challenging to optimize models for latency-critical edge applications, thus a direct DNN optimization strategy for latency is more critical to efficient edge intelligence systems.

\begin{figure}[t]
    \centering
    \includegraphics[width=0.48\textwidth]{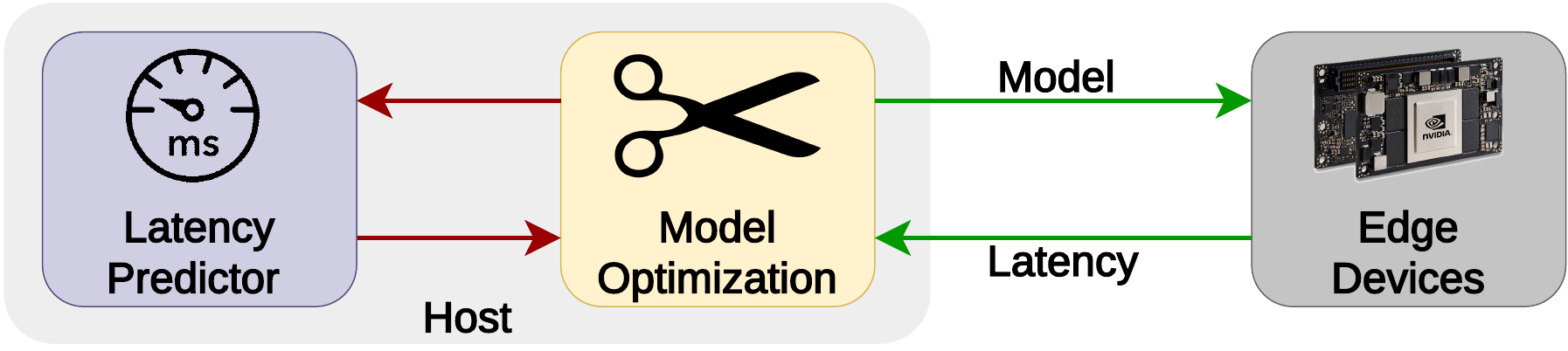} 
    \caption{The workflow of hardware-aware neural network optimization. With a latency predictor, the host can avoid time-consuming on-device measurements during the process, extremely improving efficiency.}
    \label{fig:framework}
\end{figure}

Secondly, when considering latency as the optimization objective, these methods need to deploy models to the device for measurement after making any modifications throughout the whole optimization process. As demonstrated in Fig. \ref{fig:framework} (green lines), this measurement process is quite time-consuming because the interaction with hardware to obtain the latency usually takes minutes \cite{ChamNet}, which is prohibitively expensive in a large design space. Furthermore, the indirect performance targets (e.g., FLOPs) used by these methods \cite{NetworkSlimming, SoftPrune,FPGM,PGMPF, oto, tan2019efficientnet} must be adjusted iteratively to satisfy the latency requirement upon the target edge system. This aggravates the overhead issue. 

Thirdly, some methods are very time-consuming and are mainly divided into three parts: pre-training, optimizing, and re-training \cite{NetworkSlimming}. A pre-trained model is needed to identify the importance of different parts of the model, which is used for optimizing procedures to eliminate or add parameters. Re-training fine-tunes the new model to recover accuracy. As the training procedure is complicated and time-consuming, the high overhead of network optimization is mostly due to the procedures of pre-training and re-training. Meanwhile, in order to derive a model that meets the optimization target, some prior methods \cite{NetworkSlimming, Han_weightpruning} must do the optimizing-retraining procedure for several rounds. This further exacerbates the overhead issue. 

However, previous work \cite{Rethinking} has demonstrated that pre-training is not a necessary step to obtain an efficient and competitive model. It has also stated that the architecture,  rather than the inherited weight value of significant elements, is more crucial to the efficiency of the final model. Therefore, an efficient learning method, which directly optimizes the model architecture without a pre-trained model and re-training process, is necessary to reduce the time cost.

This paper aims to develop a comprehensive and hardware-aware DNN optimization framework which effectively and efficiently fits various DNN models on diverse edge systems. We propose a neural network optimizing framework to obtain as high accuracy as possible under the latency constraint by only one learning process. We first design a compact learning scheme, which can condense redundant DNNs to meet a `hard' latency requirement on targeted devices by dynamically zeroizing and recovering Batch Normalization (BN) layers. Based on it, we propose a scaling scheme for simple models to improve their accuracy and avoid the expensive scaling factor search process \cite{tan2019efficientnet}. Thus, this framework is a comprehensive and adaptive learning method that can optimize both complex and simple models. The primary contributions of this paper are as follows:



\begin{itemize}

    \item We introduce a compact model learning method that can extract the optimal DNN architecture to fulfil the `hard' latency constraint and maintain high accuracy. It applies only a \textbf{one-shot} training procedure (named compact learning) to avoid the expensive pre-training/re-training cost. The global importance rank of channels and the dynamic Zero-Recovery process are proposed to extend the exploration space for better architecture. \textbf{(see Section \ref{section: pruning})}
    

    \item We present a machine learning-based latency predictor, which is embedded into our adaptive learning framework to provide a guideline for each Zero training phase so as to avoid time-expensive on-device measurements. This method can be also generalized to build other hardware metrics predictors.  \textbf{(see Section \ref{section:predictor})}

    \item We involve a model scaling scheme to better deploy simple models on high-end edge devices, thereby improving the accuracy and still meeting the latency constraint. Benefiting from our compact learning can optimize models from both channel and layer, the scaling method does not require the time-consuming search process for scaling factors. \textbf{(see Section \ref{section: scaling})}

    \item We demonstrate that our adaptive learning framework can effectively and efficiently optimize redundant or simple models to suit various edge devices under the specific latency constraint through a one-shot training process by numerous experiments, surpassing previous methods in terms of both accuracy and efficiency.  \textbf{(see Section \ref{section:experiment})}

\end{itemize}

The rest of this paper is organized as follows. Section \ref{section:relatedwork} introduces the background and some related works. Section \ref{section:methodology} describes the details of our proposed method, and the experimental results are presented in Section \ref{section:experiment}. Finally, we conclude this paper in Section \ref{section:conclusion}.

\section{Related Work \& Background}
\label{section:relatedwork}

In this section, we introduce the preliminaries of model optimization and latency prediction, including achievements and some flaws of these related works.

\subsection{Neural Network Compression}

Many researchers have proposed methods for reducing the high complexity of current complicated DNN models to suit the emerging embedded systems. These methods include neural network pruning \cite{NetworkSlimming} -- reducing the number of weights in a model, quantization \cite{zhao2019improving} -- reducing the number of weight bits in a model, knowledge distillation \cite{kd} -- modifying the model's architecture to simplify it. 

\subsubsection{Neural Network Pruning}
Recent efforts have been made either on unstructured pruning (i.e., weight pruning)  \cite{Han_weightpruning,louizos2018learning} or structured pruning (i.e., channel/layer pruning) \cite{NetworkSlimming,SoftPrune, FPGM}. Unstructured pruning methods directly delete individual unimportant elements in weight tensors, which can achieve less accuracy loss. However, they need specific hardware and libraries that support sparse to run pruned models.  In contrast, structured pruning methods are coarse-grained, which remove entire regular regions (e.g., channel/layer) of weight tensors, no extra supports are required. Thus, structure pruning methods are easy to deploy on different kinds of processors. Since this paper aims to optimize models for universal edge devices, we design an efficient compact model learning framework by eliminating channels and layers. As shown in Fig. \ref{fig:depth-width}, removing layers or channels can reduce the latency. However, the latency reduction of different model architectures has different tendencies to layer pruning or channel pruning under the same FLOPs. Thus, we reasonably include both channel pruning and layer pruning into our framework to determine the optimal approach through the learning process.

Conventional pruning strategies include a typical 3-stage process: 1) \textit{pre-training}, 2) \textit{pruning}, 3) \textit{re-training}. However, the pre-training stage is unnecessary to get the efficient final compact model \cite{Rethinking}. Meanwhile, it is the model architecture, rather than the inherited weight value of essential elements, that is more crucial to the efficiency and accuracy of the final model \cite{Rethinking}. Therefore, this paper aims to propose an efficient method to find the optimal compact architecture without the pre-trained model and re-training step.
\textbf{\begin{figure}[!t]
    \centering
    \includegraphics[width=0.48\textwidth]{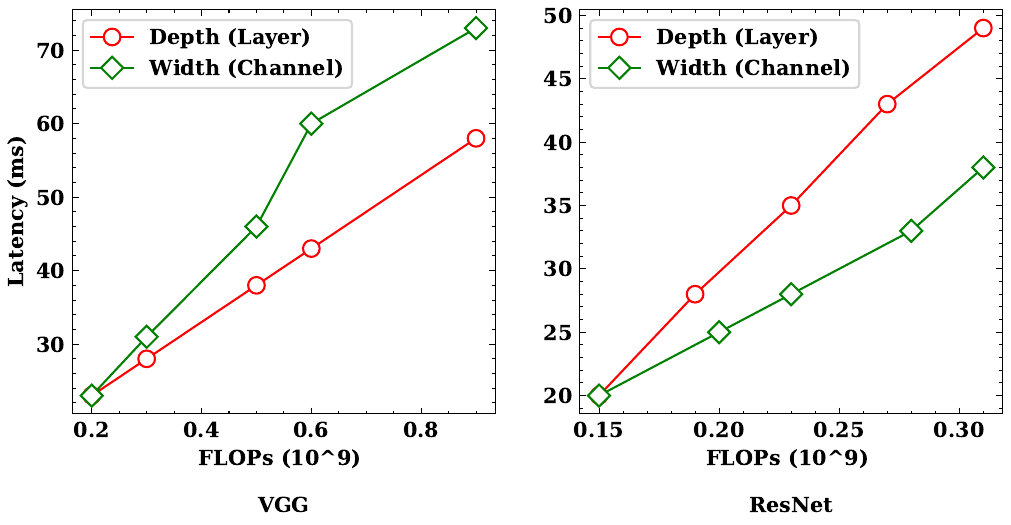}
    \caption{
The relationship between the number of FLOPs and the inference latency by changing the number of channels (width) or layers (depth) measured on NVIDIA Jetson TX2.}
    \label{fig:depth-width}
\end{figure}}

To obtain an efficient model architecture, we should rank the channel importance across layers (i.e., global channel importance ranking) to preserve important channels throughout the whole model, rather than pre-defining pruning ratios of each layer. Among global channel importance ranking methods, using batch normalization (BN) \cite{bjorck2018understanding} as the scaling factor is more efficient. Network slimming (NS) \cite{NetworkSlimming} leverages BN to remove the redundant channels and achieves some good results. However, there are some limitations within this work. First, it employs a `\textit{hard}' method to remove channels, which means that if the channels are eliminated during the pruning process, they will not be recovered in the afterwards training process. Another drawback is that after the pre-training step, the importance rank of all channels is fixed,  implying that the compact model obtained from a large compression ratio must be a sub-model of a small compression ratio. The fact is there are many different settings for each layer's channel number, hence their method blinds the design space to obtain a more efficient architecture for the specific compression ratio. In this paper, we aim to design a dynamic method to make the compression decision. 


Soft Filter Pruning (SFP) \cite{SoftPrune}, similar to our method, exploits a learning scheme in which some unimportant channels are set to zero according to the $L_p$ norm. Unfortunately, this method uses local (intra-layer-wise) channel importance ranking, so they need to initialize the pruning ratio for each layer manually by experience. Thus, this method only tries to find the `significant' weights but not the `optimal' architecture. However, previous work \cite{Rethinking} has demonstrated that the pruned architecture is more crucial to the efficiency of the final model, rather than the inherited weight value of important elements. Furthermore, it is observed that in SFP, zeroized filters are difficult to be recovered. Meanwhile, SFP does not have an initial training process so it cannot effectively discriminate the channel importance, resulting in low accuracy (similar to $s = 0$ in Section \ref{autoparaset}). The flaws of NS and SFP lead to more accuracy loss of the compact model seen in Section \ref{section:experiment}.

\textcolor{crimson}{Some existing works also compact models by one training process without the typical 3-stage process, for example, Prior Gradient Mask Guided Pruning-Aware Fine-Tuning (PGMPF) \cite{PGMPF}, which selectively suppresses the gradient of those ”unimportant” parameters via a prior gradient mask generated by the pruning criterion. It elegantly unifies pruning and fine-tuning. However, the training time overhead of PGMPF is huge, even larger than that of methods with pre-training and re-training, as shown in Section \ref{overhead}. Thus, it is also inefficient in finding the optimal compact model architecture. }

\subsubsection{Neural Network Quantization}
Quantization is used to map a large set of input to a smaller set of output. In DNN model quantization, instead of adopting the 32-bit floating point (FP-32) format to represent weights, quantized models use more compact formats such as integers or 16-bit float point (FP-16) format for weights. Existing methods on DNN model quantization mainly include training from scratch \cite{choi2018bridging, wu2018training}, fine-tuning from a full-precision model \cite{zhou2017incremental,han2015deep} or post-training quantization \cite{sung2015resiliency, shin2016fixed}. The first two methods introduce another training during the quantization process and these methods reduce accuracy drop significantly only when quantized to very low bits (e.g., below 4-bits). Meanwhile, post-training methods have already been used in commercial scenarios and perform very well in common-used 16/8-bits, like NVIDIA’s TensorRT \cite{migacz20178}. As these post-training methods do not bring extra operations in the training stage and have enough accuracy for commercial usage, we can adopt the post-training methods \cite{zhao2019improving} to obtain an efficient framework in this paper. After quantization, we can reduce the latency of a model, which indicates the quantized model can keep more channels or layers than the full-precision one under the same specific latency constraint, hence the quantized model owns higher accuracy.

\subsubsection{Knowledge distillation}
Knowledge distillation \cite{kd} is also promising to design compact models.  However, the compact model (student) has to be designed manually, whereas our approach is an end-to-end method to automatically learn a compact model from the redundant model (teacher). In future work, we are like to explore combining knowledge distillation and our method to derive more competitive models.

\subsection{Neural Network Scaling}

Model scaling \cite{tan2019efficientnet} will increase the model complexity to improve the prediction accuracy and hardware utilization, and we can use it for simple models to obtain higher accuracy and still meet the latency constraint. Model scaling can scale a model up from three dimensions, including the depth, the width and the resolution. Scaling network depth means inserting more layers into the model, scaling network width means expend channels in some layers of the model and scaling the resolution means applying the input images with a higher resolution. EfficientNet \cite{tan2019efficientnet} has demonstrated the results of scaling combined with these three dimensions. They use an expensive search process to find scaling factors for these three dimensions respectively to determine the expanded amount of each dimension for an efficient architecture and high accuracy. And they find that larger resolutions can get higher predicted accuracy, but the accuracy improvement is limited without contributing to the model architecture. Previous works \cite{raghu2017expressive} also show that the depth and width of a model are both important for high accuracy, and Fig. \ref{fig:depth-width} illustrates that scaling depth and width can lead to different increase trends of latency for different models. In the paper, we introduce model width scaling and depth scaling into our framework to  improve the utilization of simple models on edge devices for higher accuracy. Our scaling method is based on our compact learning, and we can also get the scaled model with only one training process without pre-training and re-training steps. Also, we do not need the expensive pre-searched depth and width scaling factors \cite{tan2019efficientnet, kong2022hacscale}.

\subsection{Latency Prediction}
To speed up inference, previous model optimizing works focus on changing FLOPs or channels/layers of models. However, these metrics cannot reflect the actual latency. Optimization based on real latency can better explore hardware features and hence offer additional advantages \cite{yang2018netadapt}. Only a relatively small amount of work has been done to design model compression/scaling methods considering latency as the direct optimization goal. Currently, there are some works using look-up tables (LUTs) \cite{ChamNet,yang2018netadapt,FBNet} or hardware simulators \cite{gholami2018squeezenext} to predict the latency of neural networks on devices, and they have achieved good works. Yang et al. \cite{yang2018netadapt} build layer-wise LUTs with pre-measured latency of each layer and sum up the layer-wise results to estimate the network-wise latency. Dai et al. \cite{ChamNet} and Wu et al. \cite{FBNet} construct operator-level latency LUTs by measuring the latency of each operator on real devices and computing the overall latency of models by adding up the latency of each operator in the corresponding model. 

Nonetheless, a LUT can only predict the latency of some models/layers whose structures are already in its table, hence it is only suitable for that all structures are known, and it needs rebuilding when a new structure is added. During the learning process, the number of input channels and output channels of each layer and other size-related parameters will be changed continuously. Thus, the LUT method is not available. And unlike FPGA, most hardware is a black box to users, hence it is very hard to simulate hardware by analysing its resource and scheduling algorithms, indicating that the hardware simulator is also unavailable. 
Therefore, a universal hardware latency predictor is essential to guide the learning algorithm to find a better architecture for a specific edge device. Justus et al. \cite{justus2018predicting} proposed a latency predictor based on machine learning, which can predict the execution time of the whole model. But its training data is randomly set, lowering the accuracy of predicting real models. Zhang et al. \cite{zhang2021nn} proposed nn-Meter to use a more fine-grained LUT approach which characterizes models at the kernel level, but their approach involves exhaustive testing of all possible combinations of operations (kernels) which can take anywhere up to 4.4 days. But our predictor is layer-wise and operation-level which consumes much less time (a few hours).

\begin{figure}[t]
    \centering
    \includegraphics[width=0.48\textwidth]{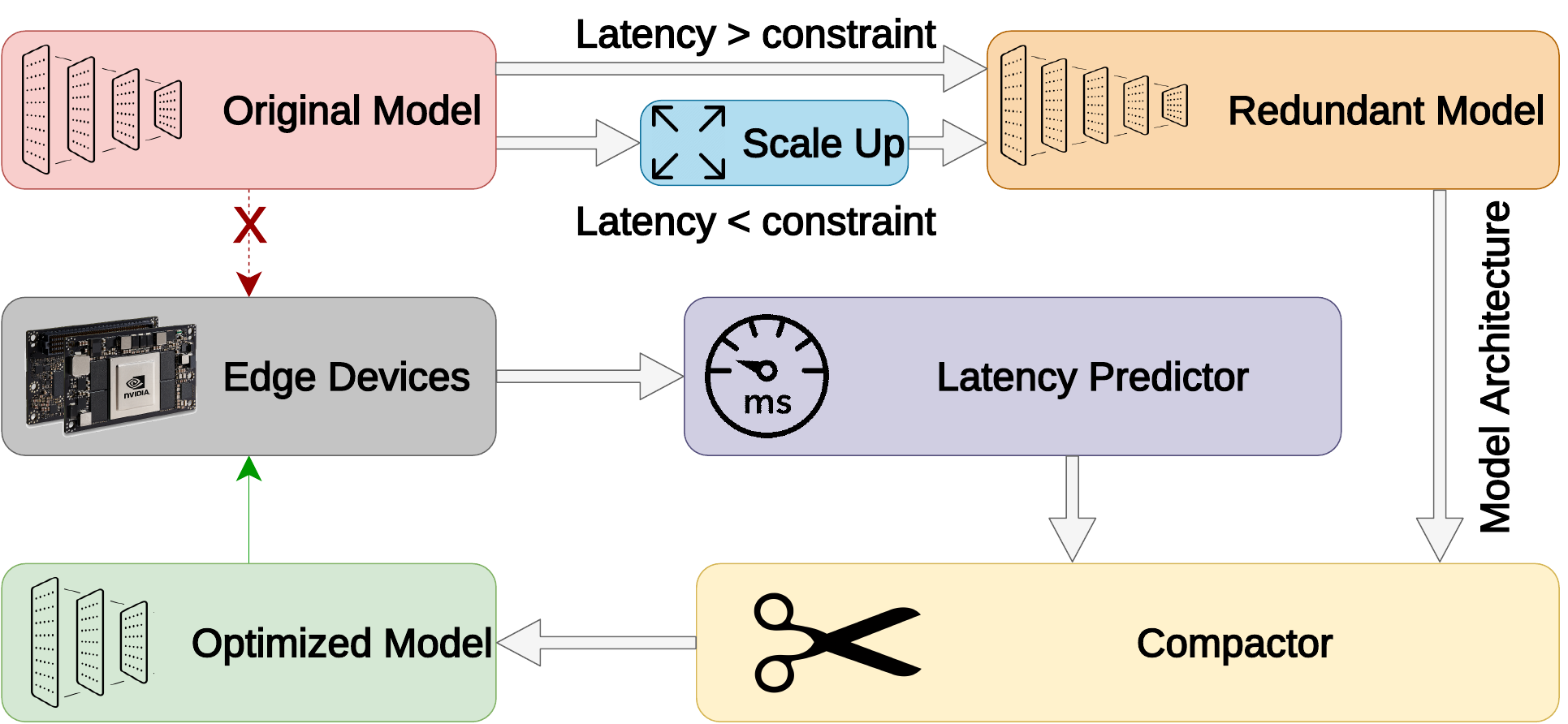}
    \caption{The overview of our model learning framework. The input is a mismatched original model and this framework outputs a model with high accuracy under the latency constraint.}
    \label{fig:zerobn-overview}
\end{figure}

\section{Latency-constraint Architecture Learning Framework}

\label{section:methodology}

In this section, we show the process of our learning framework, which trains DNNs for an edge device under a latency constraint and optimizes  complex/simple models to suit edge devices.
This framework is based on our compact learning method and it reasonably uses model scaling to expand the applicability and improve accuracy, which is shown in Fig. \ref{fig:zerobn-overview}. It generates a latency predictor from the specific edge device in advance and then executes the whole algorithm on a high-performance host machine. After evaluating the latency of the original model, if the latency is smaller than the system latency constraint, we first scale the model up to obtain a new model that exceeds the constraint. Otherwise, we keep the model unchanged. Thus, the current model is a redundant complex model that needs to be compressed to satisfy the latency constraint. For simple models, scaling up first and then compacting can eliminate the expensive scaling factors search process and improve the accuracy \footnote{\textcolor{crimson}{For overfitting considerations, see Section 3.3.}} under the latency constraint with one training process. The latency predictor also guides the compression process to find out the appropriate compact ratio for the specific latency constraint. If quantization inference is supported, we can further improve the model accuracy under the current latency constraint by integrating quantization. To use quantization, a quantized latency predictor is generated in the first step.

In the following, we first introduce the compact learning scheme, then we show how to build the universal latency predictor, and the next is our model scaling method. Finally, we introduce how to set the hyperparameters introduced by our algorithm and its further potential.

\begin{algorithm}[t]
\small
\label{code:hdsp}
\KwIn{model, training data and settings, start epoch: $s$,  zero interval: $k$, \mbox{latency constraint: $l$,} \mbox{layer pruning: $lp$.} }
\KwOut{the compact model and its parameters}

\caption{Latency-Critical Compact Learning}
\begin{algorithmic}[1] 
\STATE Initialize model and its parameters;\
\textcolor{crimson}{ \STATE Preprocess model as Fig.~\ref{fig:layer_p} if $lp$;\ }
\FOR{$e \leftarrow 1$ \textbf{to} $s-1$}
 \STATE $train\_with\_sparsity()$;\
\ENDFOR
\FOR{$e \leftarrow s$ \textbf{to} $epoch_{max}$}
 \STATE $train\_with\_sparsity()$;\
 \IF{$e \% k == s \% k$ or $e == epoch_{max}$}
 \STATE $Imp\_{rank} \leftarrow $ Sort channels by $|\gamma|$;\

 \STATE $Comp\_{ratio} \leftarrow Predictor(l,Imp\_{rank},lp)$;\

 \STATE Find threshold $\gamma_{thre}$ of $\gamma$ by $Imp\_{rank}$, $Comp\_{ratio}$;\
 \STATE Zeroize $\gamma_i$ and $\beta_i$ if $|\gamma_i|$ $<$  $\gamma_{thre}$;\ 
\textcolor{crimson}{ 
   \IF { $lp$}
 \STATE Calculate the sum of importance values in each layer  $v_i = \sum |\gamma_i|$;\ 
 \STATE Remove the layer if $v_i < v_{thre}$;\ \STATE Remove the added branch if $v_i \geq v_{thre}$;\ 
      %
  \ENDIF }

 \ENDIF
\ENDFOR
\STATE Remove all channels with zero $\gamma_i$ and $\beta_i$;\
\end{algorithmic}
\end{algorithm}

\begin{figure*}[t]
    \centering
    \includegraphics[width=0.98\textwidth]{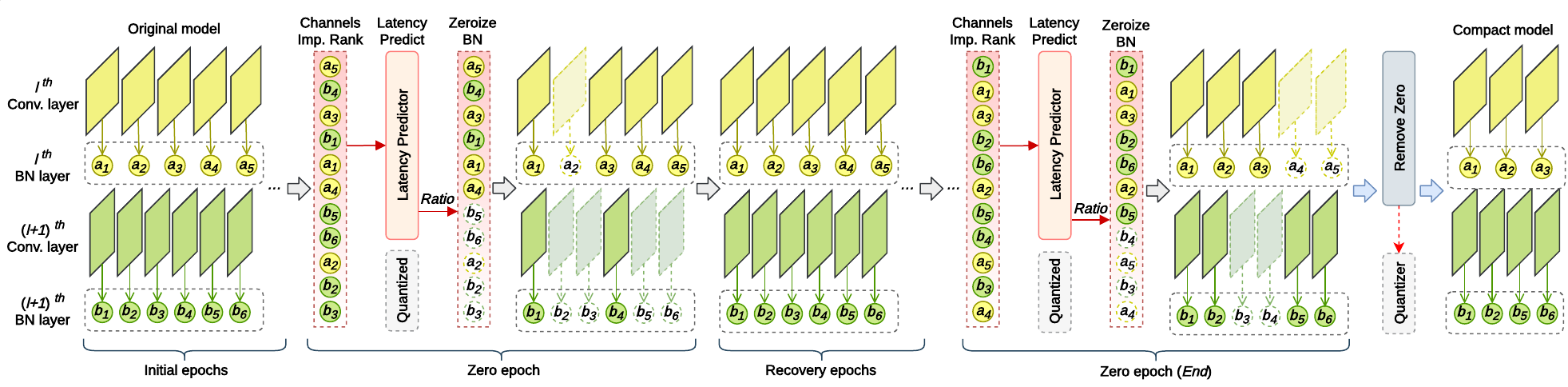}
    \caption{The process of our compact learning approach. The input is a redundant model and the output is a compact model that meets the system latency constraint. The gray dotted boxes represent the process of quantization. Each oblique rectangle represents a channel.
    }
    \label{fig:zerobn}
\end{figure*}

\subsection{Latency-Critical Compact Learning}
\label{section: pruning}

Traditional neural network pruning procedures include 3 stages:  \textit{pre-training}, \textit{pruning}, and \textit{re-training}. The pre-trained over-parameterized model, however, is not necessary to obtain an effective compact model, and it is the network architecture that is more important to the efficiency and accuracy of the final compact model \cite{Rethinking}. Thus, the compression approach presented in this paper intends to directly learn an efficient compact architecture under the specified latency constraint during the training stage of the seed model, avoiding the time-consuming three steps. The key steps in our compact learning technique are depicted in Fig. \ref{fig:zerobn}, where the input seed model is a large redundant DNN model and the output is a compact model that meets the latency constraint. The seed model does not need to be well-trained, which means that all we need is only the architecture of the DNN model being worked with. Finally, the framework will produce a trained compact model with high accuracy under the latency constraint.

In the latency-critical compact model learning process, there are two critical challenges: one is determining an efficient compact architecture extracted from the original model, and the other is determining the compression ratio under a latency constraint. To address the first challenge, We present a novel technique to automatically cultivate an efficient sub-model architecture while training the seed model from scratch.  For the second problem, we design a machine learning-based latency predictor to directly determine the number of channels or layers to eliminate. As illustrated in Algorithm \ref{code:hdsp}, the proposed compression approach only requires one training procedure, mainly including 3 phases: 1) \textbf{Initial Training}; 2) \textbf{Zero Training}; 3) \textbf{Recovery Training}. \textcolor{crimson}{The input of this algorithm is \textit{model} -- the model architecture without trained parameters (weights); \textit{training data and settings} -- the training dataset and training settings, like epochs, learning rate, and so on; \textit{start epoch} -- the epoch at which starts {Zero training}; \textit{zero interval} -- the epoch interval between {Zero training} and {Recovery training}; \textit{latency constraint} -- the latency constraint that the model being trained needs to satisfy; \textit{layer pruning} -- whether layer pruning is used in the training process. More details regarding these inputs are introduced in the following sections. The output of this algorithm is a well-trained model that satisfies the latency constraint.}

\subsubsection{Initial Training}

Initial training  (\textit{Line 3-5}) is the first step of our compact learning process and it is similar to that of conventional DNN model training, with the goal of obtaining trained weights for the seed model. In order to improve the accuracy and efficiency of the final compact model, we need an explicit learning medium to find out the optimal architecture, such as the number of layers in this model and the number of channels of each layer. According to previous work \cite{huai2021zerobn}, we can leverage the trainable parameter $\gamma$ of the BN layer to offer useful instruction for compact architecture learning without incurring any overhead. As shown in Eq. (\ref{eq-bn}), $x$ is the input of the BN layer and $y$ is the output.  $\epsilon$ is a number that is artificially set to keep the denominator from becoming zero. The mean and standard deviation of a mini-batch $B$ are $\mu_{B}$ and $\sigma_{B}$, respectively. $\gamma$ and $\beta$ are affine transformation parameters that may be trained to translate normalized activation to any scale linearly. Since each channel has a corresponding $\gamma$, they can be utilized as explicit medium directly to reflect the importance of channels. These $\gamma$ is collaboratively optimized with the network weights during the training phase, thus, the network can automatically recognize the importance of each channel, and channels with small $\gamma$ can be safely removed.


\begin{equation}
\label{eq-bn}
    \overline{x} = \frac{x - \mu_{B}}{\sqrt{{\sigma^{2}_{B}}+\epsilon}}; \quad y = \gamma \cdot \overline{x} + \beta
\end{equation}

To determine the optimal architecture of the compact model, the seed model should be trained for several epochs, allowing all explicit medium $\gamma$ to have informative values that represent the importance of channels (\textit{Line 3-5}) rather than randomly initialized numbers (\textit{Line 1}). Meanwhile, the Initial training process also guarantees that the updating directions are not 0 (used in Recovery training). \textcolor{crimson}{We use sparsity training \cite{NetworkSlimming} throughout the training phase to impose sparsity regularization on these explicit medium $\gamma$. Although sparsity training was also applied in the pre-training phase of NS's \cite{NetworkSlimming}, our initial training requires fewer epochs, as shown in Section \ref{section:experiment}. Meanwhile, compared to the pre-training stage in other works \cite{li2016pruning, molchanov2016pruning}, which simply apply traditional DNN model training, the sparsity training used in our initial training can better distinguish the importance of these explicit medium $\gamma$ for better compact DNN architecture.} After Initial training epochs, we can formally enter the two most essential iterative phases of our method: Zero and Recovery (\textit{Line 6-19}). Our approach can extract an optimal compact architecture by dynamically adjusting it according to the trained $\gamma$ during these two iterative phases.


\subsubsection{Zero Training}

Zero training (\textit{Line 6-19}) is used to compress the model and extract a compact architecture. We sort all the $\gamma$ according to their absolute value first to obtain the global importance rank of channels across all layers  (\textit{Line 9}). Then our proposed latency predictor reports a compression ratio to satisfy the specified latency constraint (\textit{Line 10}). The details about our machine learning-based latency predictor are described in Section \ref{section:predictor}. The compression ratio and the global importance rank are used to determine the threshold of $\gamma_i$ (\textit{Line 11}). Then, for insignificant channel $c_i$ whose $\gamma_i$ is less than the threshold, we zeroize out those corresponding $\gamma_i$ and $\beta_i$ (\textit{Line 12}) to let these channels in BN layers output zero. When the BN layer is applied after a convolutional layer, each channel of the BN layer corresponds to one channel of the previous layer, implying that the Zero process is equivalent to removing those channels from the convolutional layer. Therefore, our compression method has extracted a compact architecture through the Zero process to fulfil the certain latency constraint. Meanwhile, it retains all of the trained weights of convolutional layers in this model and only simply zeroizes out the associated $\gamma_i$ and $\beta_i$. Each channel contains a $\gamma_i$ value and a $\beta_i$ value, thus, this  Zero training just removes a small number of values from the trained model, making it easier to recover.

\begin{figure}[!t]
    \centering
    \includegraphics[width=0.48\textwidth]{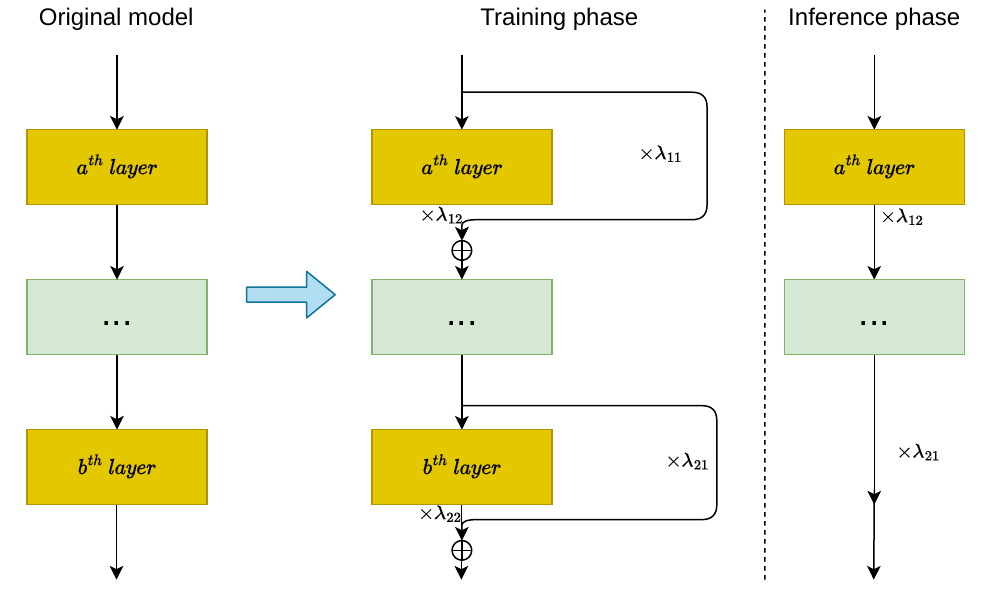}
    \caption{The preprocessing for layer pruning by adding branch layers.}
    \label{fig:layer_p}
\end{figure}

\textcolor{crimson}{\textbf{Layer Pruning:}} As shown in Fig. \ref{fig:depth-width}, different model architectures have different latency trends by changing the number of channels or layers. Thus, in order to derive the best model architecture, layer pruning is as important as channel pruning. However, eliminating layers directly during the training process results in a disconnected model architecture. Thus, for supporting layer pruning,  before training the model, we need to pre-process it (\textit{Line 2}). As illustrated in Fig. \ref{fig:layer_p}, we add a directly-connected layer as a branch to each convolutional layer and we aggregate the results from the convolutional layer and its associated branch layer to get the input of the next layer \footnote{If the output size is different to that of the branch layer, only aggregate outputs in corresponding positions.}. In the Zero training,  the latency predictor only needs to predict one branch latency as we only retain one branch for inference. \textcolor{crimson}{To eliminate layers, after zeroizing $\gamma_i$ and $\beta_i$ corresponding to unimportant channel $c_i$ in each layer (\textit{Line 12}), we calculate the sum of each layer's channel importance values and remove a layer (zeroize the corresponding $\lambda$) if its total importance values are less than the threshold (\textit{Line 14-15}).  Otherwise, we remove the corresponding added branch (\textit{Line 16}). The larger the important value, the more ``information'' the corresponding channel retains. And the sum of important values in a layer means the ``total information'' this layer keeps. In the layer pruning, we only eliminate those layers with tiny total important values to maintain more ``information'' for high accuracy. Thus, during the compact training, if the sum of important values in each layer is larger than the threshold $v_{thre}$, we do not remove any layer in this model, implying that layer pruning does not occur in this training process. This is why some models did not have the experimental result of layer pruning in Section \ref{section:experiment}. }



\subsubsection{Recovery Training}
After the Zero training phase, the Recovery training process (\textit{Line 6-19}) is executed for $k-1$ epochs to allow the zeroized parameters (\textit{Line 8}), such as $\gamma_i$ and $\beta_i$, to recover themselves rather than being permanently removed. If there are channels that are eliminated in the previous Zero training phase but are potentially essential, the Recovery training can help them jump out of zero and play a crucial part in the subsequent training process and even the inference processes. The Recovery training gives our method a chance to learn more efficient architecture with higher accuracy. Eq. (\ref{eq-zerobn-1}) indicates the common calculation formula in a convolutional layer with BN and activation function, where $L$ is the index of layer, $A^L$ is the output of layer $L$, $W^L$ is the weight between layer $L-1$ and $L$, and $\sigma^L$ is the activation function at layer $L$. $\ast$ represents the convolution operation and other parameters are the same as Eq. (\ref{eq-bn}). 
\begin{equation}
\label{eq-zerobn-1}
 A^L = \sigma^L(\gamma^{L}( \overline{A^{L-1} \ast W^L}) + \beta^{L})
 \end{equation}
In the Recovery training, we focus more on updating $\gamma^{L}$ and $\beta^{L}$ that are zeroized during the Zero training (the $\lambda$ used in layer pruning is similar). Eq. (\ref{eq-zerobn-2})-(\ref{eq-zerobn-3}) show the gradients of $\gamma^{L}$ and $\beta^{L}$ to the final loss function.

\begin{equation}
\label{eq-zerobn-2}
   \frac{\partial loss}{\partial \gamma^L } = \frac{\partial loss}{\partial A^L}\cdot {\sigma^L}^{'} \cdot ( \overline{A^{L-1} \ast W^L})
 \end{equation}
\begin{equation}
\label{eq-zerobn-3}
   \frac{\partial loss}{\partial \beta^L} = \frac{\partial loss}{\partial A^L}\cdot {\sigma^L}^{'}
 \end{equation}
Since we have zeroized some values in $\gamma^{L}$ and $\beta^{L}$, and the most often used activation function is \textit{ReLU}, the corresponding output of $\sigma^L$ is likewise zero under \textit{ReLU}. Although \textit{ReLU} is not differentiable at zero, it is widely accepted that its derivative is also zero. Thus, we can easily derive Eq. (\ref{eq-zerobn-4}).
\begin{equation}
\label{eq-zerobn-4}
  \gamma^L, \beta^L = 0 \rightarrow {\sigma^L}^{'} = 0 \rightarrow \frac{\partial loss}{\partial \gamma^L }, \frac{\partial loss}{\partial \beta^L} = 0
\end{equation}
Until now, we may need to offer these zeroized $\gamma^{L}$ and $\beta^{L}$ some small values to break this deadlock. However, most training algorithms employ momentum in the optimizer to speed up convergence, as the momentum accumulates the gradients of the past steps to determine the direction to go, rather than using only the gradient of the current step \cite{mm}. Eq. (\ref{eq-zerobn-5})-(\ref{eq-zerobn-6}) illustrates the updating rules of weight with momentum, in which $lr$ is the learning rate, $z_k$ is the updated value of the last step and $m$ is the accumulation coefficient. Following these rules, $w_k$ are updated to $w_{k+1}$ by combining current and past gradients.
\begin{equation}
\label{eq-zerobn-5}
    z_{k+1} = m\cdot z_k +  \frac{\partial loss}{\partial w_k}
\end{equation}
\begin{equation}
\label{eq-zerobn-6}
    w_{k+1} = w_k - lr \cdot z_{k+1}	
\end{equation}
As a result, even though the current gradients of $\gamma^{L}$ and $\beta^{L}$ are zero, they can be updated to recover from zero by following the last directions (This is also why we need the Initial training). This process is fully automated and efficient, as no additional steps are required.
\begin{equation}
\label{eq-zerobn-7}
     w_k,  \frac{\partial loss}{\partial w_k} = 0 \rightarrow w_{k+1} = -lr \cdot m\cdot z_k
\end{equation}

\begin{figure}[!t]
    \centering
    \includegraphics[width=0.48\textwidth]{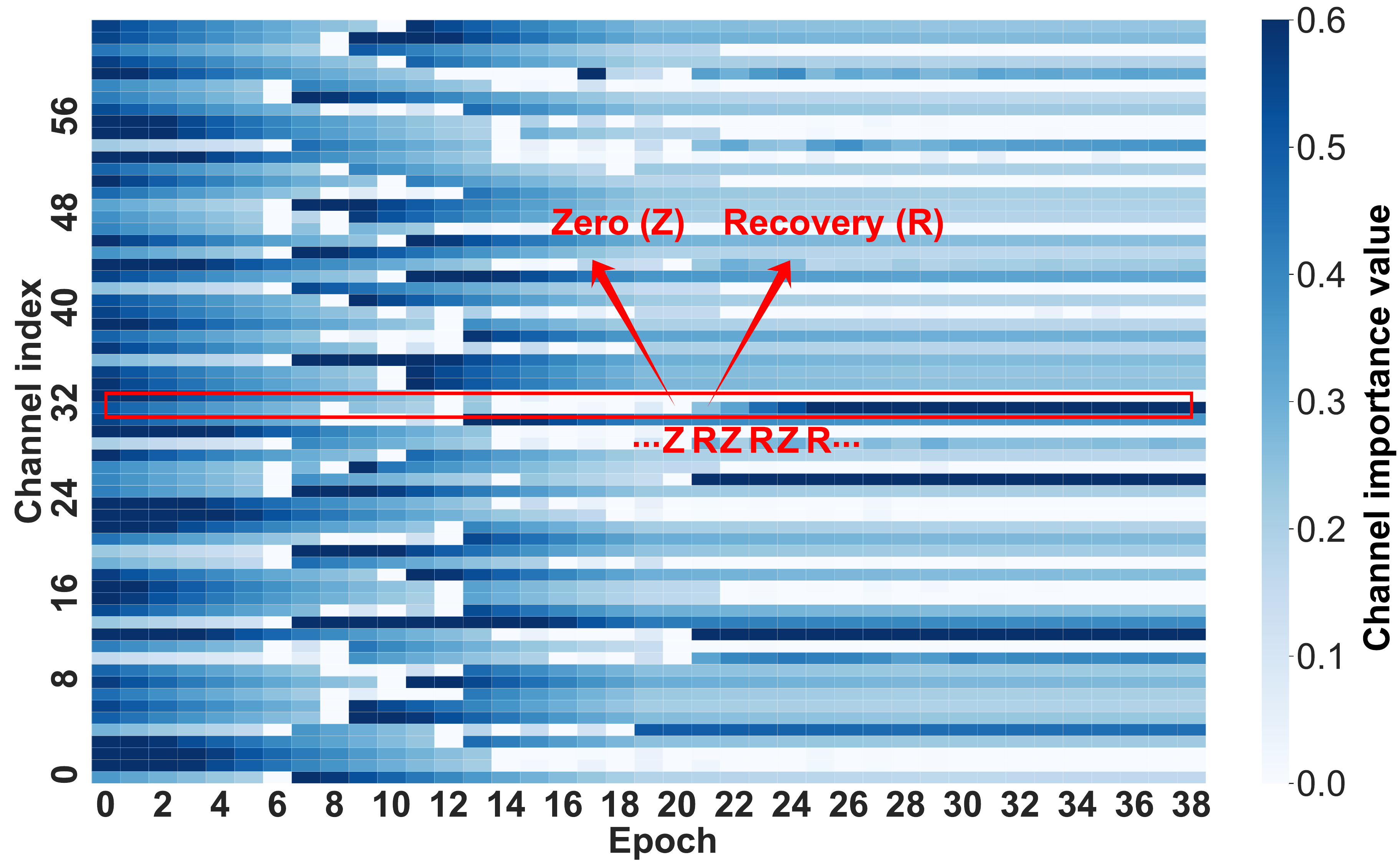}
    \caption{Channel importance changes during our learning process. Zeroized channels (white blocks) can be easily recovered.}
    \label{fig:zerobn_channel}
\end{figure}

Specifically, as the previous training phase is Zero training, the updating rule of $w_{k+1}$ is shown in Eq. \ref{eq-zerobn-7}, implying that $w_{k+1}$ are automatically set to non-zero small values ($lr$ and $ z_k$ are small). In the Recovery training, the network first  tries to predict results under those $\gamma^{L}$ and $\beta^{L}$ are small. And channels corresponding to these small $\gamma^{L}$ and $\beta^{L}$ have little effect on final results. Just like knowledge distillation, the training process tries to use a small network architecture to cover all the knowledge learned by the large network architecture. There are two situations in this process: 1) if the loss is small, then the network can exclude these channels safely for finishing compression; 2) if the loss is huge, then the optimizer updates these parameters significantly (see  Eq. \ref{eq-zerobn-5} and Eq. \ref{eq-zerobn-6} ). In the second condition, these channels zeroized in previous Zero training may become significant and we will not zeroize them in the future Zero training process, therefore the architecture of the final compact model is changed. Especially, as the architecture changes, the latency predictor will report a different compression ratio in the next Zero training phase. The training process tries to minimize loss and improve accuracy, allowing the compact architecture to improve within the specific latency constraint. Compared to \cite{NetworkSlimming}, our method can identify a more efficient and accurate model with better architecture,  as shown in Section \ref{section:experiment}. Furthermore, we do not need to retrain the compact model further as our learning process already obtains high accuracy, reducing the training overhead.


\begin{figure}[!t]
    \centering
    \includegraphics[width=0.49\textwidth]{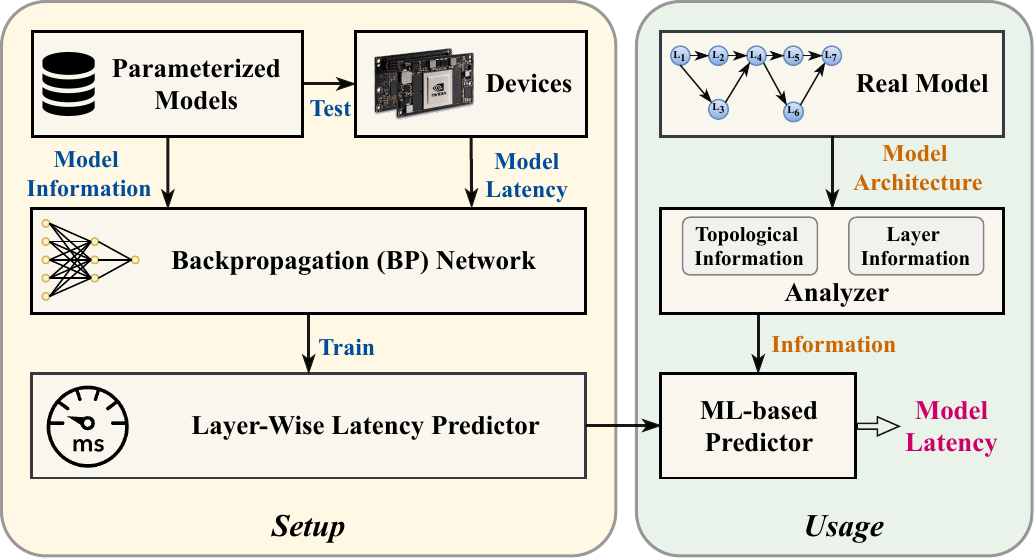}
    \caption{The framework of our latency predictor. The left process represents the building of the predictor and the right part reveals how to predict the latency of a real model with the predictor.}
    \label{fig:predictor}
\end{figure}

Fig. \ref{fig:zerobn_channel} shows the changes in the importance of all channels during the learning process on an example model, where the start epoch of Zero training $s$ is $6$, the interval epochs of Zero training $k$ is 2, and the compression ratio is set to a constant $0.3$. We can derive from this figure that zeroized channels are easy to recover and even some channels, like those framed in the red rectangle, may become very essential. We also discovered from our experiments that if we zeroize filter weights, which are utilized in SFP \cite{SoftPrune}, it is difficult for these filters to avoid being zeroized in the next pruning step by employing momentum. The reason may be gradients are also zero and each filter has a large number of weights, then the momentum is a little small for a fully zeroized filter to win other non-zeroized filters by $L_p$ norm. Thus, zerorizing $\gamma$ and $\beta$ in the BN layer is an effective way to cultivate the optimal sub-model. This is why our accuracy is higher than that of other methods.
%

After some repetitions of Zero and Recovery training epochs, removing all channels corresponding to zeroized $\gamma_i$ from the channel-sparsity model (\textit{Line 20}), we can obtain an efficient compact architecture fulfilling the specific system latency constraint. Also, we remove layers (branches) with zeroized $\lambda$. This model keeps weights from the training process, hence it does not need to be re-trained.


\textbf{\textcolor{crimson}{Training Overhead per Epoch:}} Compared to the traditional training process, we only introduce overhead in the zeroizing process. As we remove channels in convolution layers, we mainly focus on convolution. The time complexity of convolution layers for an input is $O(\sum_{l=1}^L M_l^2K_l^2C_{l-1}C_l)$, where $L$ is the number of convolution layers, $C_l$ is the number of output channels in layer $l$, $M_l$ is the output size of layer $l$ and $K_l$ represents the kernel size of layer $l$. The zeroizing operation includes a sorting step with a time complexity of $O(nlogn), n = \sum_{l=1}^LC_l$. Generally, $logn$ is much smaller than $M_l^2K_l^2C_{l-1}$ of each layer $l$, so its complexity is much lower than the convolution layers. The latency predictor only includes hundreds of multiplication and addition operations, which is much smaller than convolution operations. Thus, the overhead of our compact training can be neglected, demonstrating the high efficiency of our method. \textcolor{crimson}{ The evaluation results of overhead are shown in Section \ref{overhead}.}

\subsection{ML-based Latency Prediction Methodology}
\label{section:predictor}

Previous researches have used hardware simulator-based \cite{gholami2018squeezenext} or look-up table (LUT)-based \cite{ChamNet, yang2018netadapt, FBNet} predictors to estimate the latency of DNN on a certain device. However, the limitations of these methods become obvious throughout the model learning process. Since most devices are black boxes to users, it is difficult to emulate hardware by analyzing its resources and scheduling algorithms. As a result, most commercial edge devices are incompatible with hardware simulators. Second, LUT calculates the overall latency of DNN by summing up the recorded latency of each layer, thus, it can only provide the latency of a model that only includes pre-defined layer structures. However, during the learning process, the number of layers and the number of input channels, output channels, and other size-related parameters of each layer will be updated continuously. It is impracticable for LUT to preserve the whole design space, and LUT needs to rebuild anytime a new layer structure is introduced. Thus, LUT is not suitable for our framework, either. There are several latency predictors based on Neural Network \cite{justus2018predicting, zhang2021nn, dudziak2020brp}, but their training data is randomly chosen and network architectures are not well designed, therefore the accuracy of these predictors is lower.

In this model learning framework, we need a predictor which can provide any model latency on the specific hardware. There are three challenges in building such a latency predictor: 1) for each operation, such as convolution, each variable parameter has tens to hundreds of common choices, hence the search space is nearly infinite; 2) no hardware design details or scheduling schemes are available for reference; and 3) this predictor is should be accurate enough, as the latency is a `hard' constraint. Therefore, only methods that can extract patterns from limited samples are suitable for our framework. As illustrated in Fig. \ref{fig:predictor}, we propose a hardware-customized latency predictor based on Backpropagation (BP) neural network \cite{bp}.

\begin{figure}[t]
    \centering
    \includegraphics[width=0.47\textwidth]{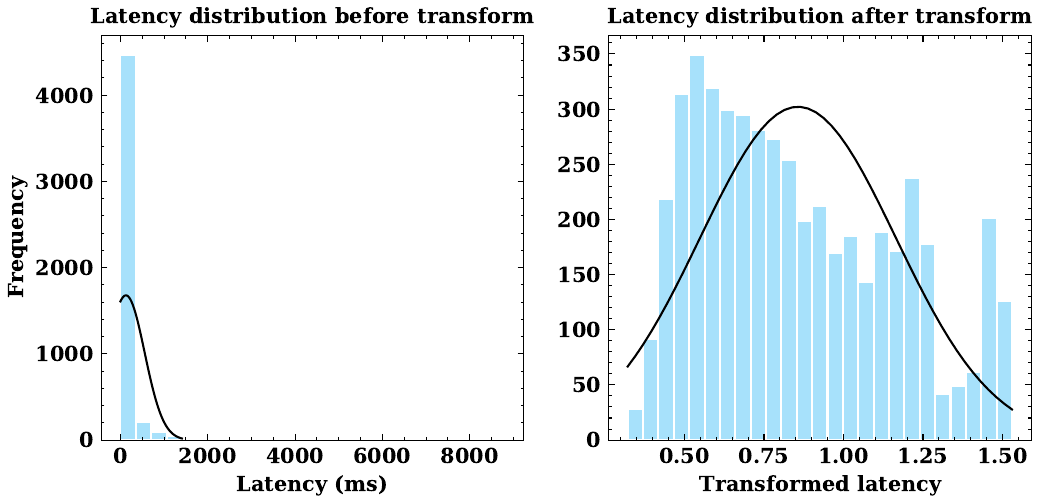}
    \caption{The latency distribution before/after transform. The normal distribution can improve the accuracy of the latency predictor.}
    \label{fig:dis}
\end{figure}

\begin{equation}
\label{bp}
    g(x) = \sigma^L(W^L\sigma^{L-1}(W^{L-1} \cdot \cdot \cdot \sigma^1(W^1x) \cdot \cdot \cdot)) 
\end{equation}
BP network is a lightweight neural network used to map the input space to the output space without expert-designed mathematical functions. Eq. \ref{bp} shows the forward propagation of the  BP network, where $x$ is the vector of features, $L$ is the number of layers, $W^l$ is the weight between layer $l-1$ and $l$, and $\sigma^l$ is the activation function at layer $l$. According to this equation, after training this BP network, hundreds of multiplication and addition operations are enough to obtain the latency, which just takes a few milliseconds, while interacting with the hardware takes several minutes. Also, it is easy to find that the updating scheme of the BP network is also simple, thus training it is not time-consuming. The BP network we used includes three layers, which are the input layer (16 neurons), the hidden layer (12 neurons), and the output layer (1 neuron). And there is an activation function between input features and every two connected layers, here we use $tansig$, $tansig$, and $purelin$, respectively.

\textcolor{crimson}{To train the BP-based latency predictor, we first need to build a dataset, which includes many parameterized models. The parameterized model refers to the single-layer model with various operations like Convolution,  BN and Pooling configured with different size-related parameters (configurations). The design space of size-related parameters is determined by the commonly-used range of each parameter. Then we undertake regular and random sampling over the whole design space of parameters. Regular sampling splits the design space into several grids, whereas random sampling supplements extra details to these grids. After constructing parameterized models with these sampled parameters, we measure them on the specific device to obtain precise latency. Then we train the BP neural network with configurations of these sampling models and their corresponding precise latency (labels), trying to let the BP network output the latency by inputting a configuration.  To improve the accuracy of the BP network, the training labels must correspond to the normal distribution \cite{de1993backpropagation}, however, the latency of most single-layer models is small. Thus, as illustrated in  Fig. \ref{fig:dis}, we need to transform these latency data labels to approximate the normal distribution before the BP network training process. } 

\textcolor{crimson}{Although the number of single-layer models in our sampling is limited, the trained BP network can predict the latency of any configuration in the design space \cite{bp}. So, the trained BP network is a single-layer predictor. To predict the overall latency of the entire model, we first use the single-layer predictor to obtain the latency of each layer in the model. Then, we analyse the model's topological information to find the latency-weighted longest path and add all latency on this path to obtain the whole model latency.}
%
Table \ref{tab:latencypredictor} shows the accuracy of the proposed BP-based latency predictor on widely-used models and the average error is about 6.12\%, whereas the error of \cite{justus2018predicting} on real models is about 13.52\%. Without loss of generality, the BP-based predictor is a common approach that is easily applied to other neural network design frameworks when considering the latency constraint. Also, this method can be also generalized to other hardware metrics.




\begin{table}[!t]
\centering
\scriptsize
\caption{Results of latency predictor on various models.}
\label{tab:latencypredictor}
   \setlength{\tabcolsep}{5pt} 
    \renewcommand{\arraystretch}{1.1} %

\begin{tabular}{ccccc}
\hline
\textbf{Method}                                                            & \textbf{Model} & \textbf{\begin{tabular}[c]{@{}c@{}}Real Latency\\ (ms)\end{tabular}} & \textbf{\begin{tabular}[c]{@{}c@{}}Predicted Latency\\ (ms)\end{tabular}} & \textbf{\begin{tabular}[c]{@{}c@{}}Error\\ (\%)\end{tabular}} \\ \hline
\multirow{5}{*}{\begin{tabular}[c]{@{}c@{}}Our \\ BP Network\end{tabular}} & ResNet-101     & 82.67                                                                & 83.19                                                                     & 0.63                                                          \\
                                                                           & MobileNet-160  & 11.67                                                                & 11.42                                                                     & 2.14                                                          \\
                                                                           & Inception-v3   & 70.60                                                                & 64.73                                                                     & 8.31                                                          \\
                                                                           & Nasnet-mobile  & 96.25                                                                & 111.15                                                                    & 13.41                                                         \\ \cline{2-5} 
                                                                           & \multicolumn{3}{c}{\textbf{Average Variation (\%)}}                                                                                                               &\textbf{ 6.12 }                                                         \\ \hline
NN  \cite{justus2018predicting}\tablefootnote{Only report results for different batch size on VGG-16.}                                                                         & VGG-16         & 65.37                                                                & 56.53                                                                     & 13.52                                                         \\ \hline
\end{tabular}
\end{table}

\subsection{Latency-critical Model Scaling}
\label{section: scaling}
For accelerating the inference, some lightweight model architectures are proposed. The latency of a simple model may be smaller than the constraint of a high-end system, indicating that this model can be extended for more accuracy but still meet the latency requirement. Previous works \cite{tan2019efficientnet,kong2022hacscale, lee2020neuralscale} have proposed some methods to scale up a model, demonstrating the efficiency of model scaling. In this work, we first scale a DNN model up uniformly (\textit{unified scaling}) and then use our compact learning method to ensure that the latency constraint is satisfied, eliminating the expensive search process of scaling factors.

\begin{figure}[t]
    \centering
    \includegraphics[width=0.47\textwidth]{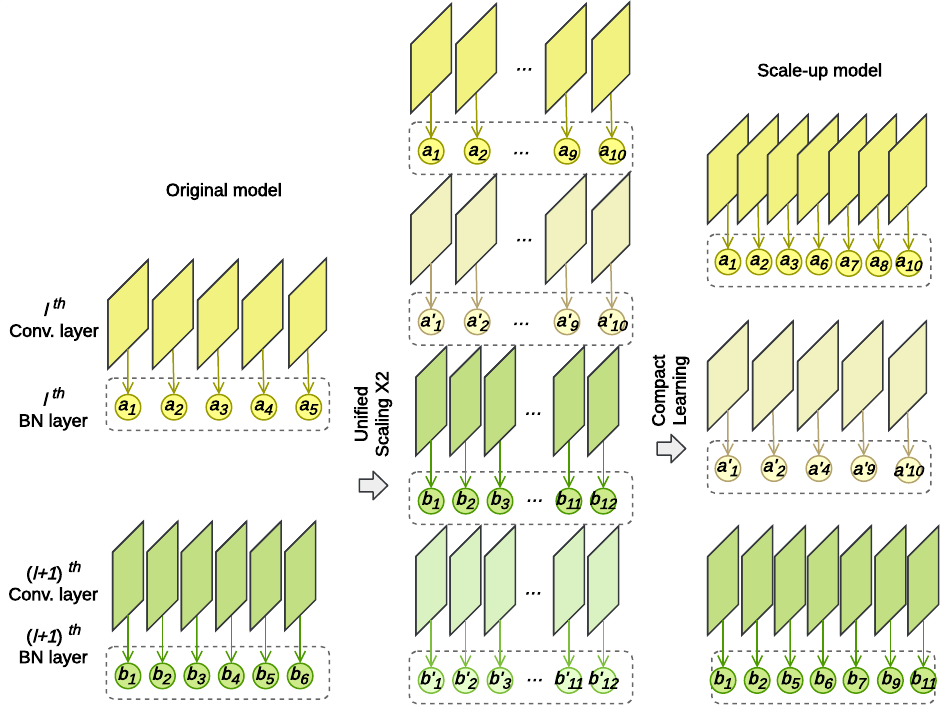}
    \caption{The process of our latency-critical model scaling approach. }
    \label{fig:scale}
\end{figure}

To undertake model scaling with our framework, we need to expand the model before feeding it into the compact learning scheme, as shown in  Fig. \ref{fig:scale}, where the input seed model is a simple DNN model and the output is an extended model with higher accuracy under the latency constraint. We also do not need to train the seed model and only its model architecture is needed. As illustrated in Algorithm \ref{code:scaling}, we first scale the model up by a factor $\delta$, which increases the number of channels in each layer and the total number of layers in the model by this factor (\textit{Line 1-2}), thus it is named ``unified scaling". The scaling factor $\delta$ is set by the latency margin from the constraint and makes sure the current latency is larger than the latency constraint. After this step, we obtain a redundant model again. Then we preprocess this model following Fig. \ref{fig:layer_p} to add branch layers (\textit{Line 3}), allowing layer pruning (\textit{Line 4}). And our compact learning can compress this extended model (\textit{Line 5}), and the final scaled-up model meets the latency requirement with higher accuracy than the original model. As our compact learning can remove channels and layers and it trains for high accuracy, we do not need the expensive search \cite{tan2019efficientnet, kong2022hacscale} to identify the optimal scaling factors for layers and channels, respectively.

\textcolor{crimson}{However, for small datasets such as CIFAR-10 \cite{cifar10}, commonly designed models may already be redundant. In this instance, model scaling cannot lead to higher accuracy; for example, the VGG-19 \cite{vgg} model with 70\% of channels removed can achieve higher accuracy than the full VGG-19 model on the Cifar-10 dataset (see Fig. 11). Then we can derive that, for redundant models, model scaling diminishes rather than increases accuracy. On the other hand, for large datasets such as ImageNet \cite{ILSVRC}, model scaling generally improves accuracy \cite{tan2019efficientnet, kong2022hacscale, bianco2018benchmark}. Thus, we can safely scale the model up for better accuracy on large datasets. Meanwhile, without loss of generality, we can evaluate and compare the accuracy of the original model and the scaled model to choose the better one.}

\begin{algorithm}[t]
\label{code:scaling}
\KwIn{model, training data and settings, start epoch: $s$,  zero interval: $k$, \mbox{latency constraint: $l$.} }
\KwOut{the scaled model and its parameters}

\caption{Latency-critical Model Scaling}
\begin{algorithmic}[1] 
\STATE Extend channels of each layer by $\delta$ times; (Width)\
\STATE Duplicate each layer by $\delta$ times; (Depth)\
\STATE Preprocess the extended model according to Fig. \ref{fig:layer_p};\
\STATE Set the value of $lp$ to \textit{True};\
\STATE Call \textit{Latency-Critical Compact Learning};\
\end{algorithmic}
\end{algorithm}

\subsection{Automatic Parameter Setting}
\label{autoparaset}

Although our framework is efficient for optimizing models to meet the latency constraint of an edge system by only one training process, it introduces two hyperparameters (i.e., $s$ and $k$ in Algorithm \ref{code:hdsp} and \ref{code:scaling}) into the training process, which brings some difficulty to developers. We will give a reference on how to determine these hyperparameters by our experimental data. First, for the $k$, as shown in the left part of Fig. \ref{fig:ks}, we can set the reference value to $2$ in most instances, which implies we should alternatively execute one Zero epoch and one Recovery epoch after Initial epochs. Second, for the $s$, according to our experiments, it is a little difficult to find the optimal value, as the optimal value is not fixed but in a range. Therefore, after analyzing the results shown in the right part of Fig. \ref{fig:ks}, the simple way is to set the reference value of the start epoch for our framework to half of all training epochs ($epoch_{max}$).

Particularly, we can conclude from Fig. \ref{fig:ks} that if $s$ is close to the optimal value, the accuracy is similar to the optimum. And we can also deduce that it prefers a smaller $s$ to a larger $s$, thus, we round down the result to 10 multiples to guarantee the $s$ value is not larger than the optimal one. The other settings related to Fig. \ref{fig:ks} will be introduced in Section \ref{section:experiment}.

\subsection{Model Quantization}

In this paper, the edge devices we used (i.e., Nvidia Jetson TX2 and Nvidia Jetson Nano) only support half-precision floating-point (FP16) quantized inference. For FP16 quantization, the post-training quantization method can achieve high accuracy and does not bring extra training overhead \cite{migacz20178}. Quantizing the trained model from full-precision floating-point (FP32) format to FP16 format can reduce the model complexity and lower latency. Under the specific latency constraint, the FP16 model can retain more weight than the FP32 model, resulting in higher accuracy. We use the TensorFlow Converter \cite{tensorflow2015-whitepaper} to transform models from FP32 format to FP16 format and finish the quantization process. Without loss of generality, other quantification methods \cite{zhao2019improving, sung2015resiliency} and other quantization bits (e.g., INT8) also suit our framework.
\subsection{Further discussion}

As shown in previous sections, our learning is an efficient and effective method to train suitable models for a variety of hardware. Given that BN  is included in most DNN models, we believe that our approach has a significant potential for learning efficient and accurate networks for the emerging edge era. Meanwhile, if a DNN model does not have the BN layer, we can simply introduce a value  $\gamma^{'}$ into each convolution layer as a mask. After training, we can remove all channels with zero $\gamma^{'}$ and multiply nonzero $\gamma^{'}$  to corresponding channel weights to obtain the final model, implying that this model does not include extra operations in the inference phase.

Meanwhile, not only the pre-verified models (e.g., state-of-the-art models) but also models that have not been validated (e.g., searched from NAS), can be fed into our framework to optimize for more suitable architectures. As NAS methods \cite{nas} are progressively replacing the hand-crafted model design, our framework can be integrated into NAS to design more hardware-efficient, competitive models. This is what we are like to investigate in the future.

Furthermore, our framework provides a method to trade-off between latency and accuracy for deploying machine learning algorithms better to black-box edge devices. Furthermore, the constraints of edge systems include not only latency but also energy, memory, and so on. Our framework is also easy to modify for optimizing these goals, for example, we can build a power predictor or a memory predictor by the BP network. Then the same framework can balance accuracy and energy or memory. This paper just takes latency as an example, but our framework is not limited to optimizing latency.

\begin{figure}[t]
    \centering
    \includegraphics[width=0.47\textwidth]{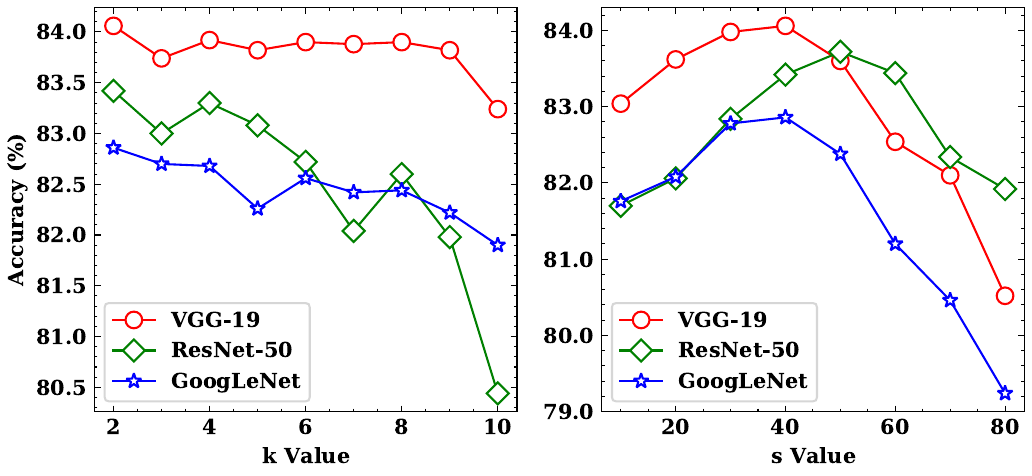}
    \caption{The accuracy w.r.t the variation of $k$ and $s$ values.}
    \label{fig:ks}
\end{figure}
\section{Experiments}
\label{section:experiment}
In this section, we first introduce the details of our experimental environment including hardware platforms, benchmark models, datasets, and training settings. \textcolor{crimson}{ Then we show the training overhead of our compact learning.} Next, we show the accuracy of our proposed learning method in comparison to state-of-the-art methods. Finally, we present the accuracy and efficiency of our framework under a hard latency constraint.

\subsection{Experimental Setup}
\textbf{Hardware Platforms:} 
The target hardware platforms we used include NVIDIA\textsuperscript{\textregistered} Jetson\textsuperscript{\texttrademark} TX2 and NVIDIA\textsuperscript{\textregistered} Jetson Nano\textsuperscript{\texttrademark}. Jetson TX2 is equipped with an NVIDIA Pascal\textsuperscript{\texttrademark} GPU with 256 CUDA\textsuperscript{\textregistered} cores and 8GB LPDDR4 memory. Jetson Nano is equipped with an NVIDIA Maxwell\textsuperscript{\texttrademark} GPU with 128 CUDA\textsuperscript{\textregistered} cores and 4GB LPDDR4 memory. These two are widely-used embedded machine learning edge devices. To maximize the performance, we put both TX2 and Nano in the ``Max-N'' mode and lock their clocks to maximums.

\textbf{Datasets:}
The datasets used for benchmark are CIFAR-10 \cite{cifar10} and ILSVRC-2012 \cite{ILSVRC}, which are commonly used for evaluating model optimization  approaches. The CIFAR-10 dataset consists of 10 classes of RGB images with a resolution of 32$\times$32. There are a total of 50K and 10K images in the training and testing sets, respectively. The ILSVRC-2012 dataset consists of RGB images in 1K classes with higher resolutions. The training set contains around 1.28M images, whereas the validation set has approximately 50K images. Following previous works \cite{imagenet100_ICCV}\cite{imagenet100_ECCV}, we randomly select 100 classes from the ILSVRC-2012 dataset to set up a subset named ImageNet-100 for more efficient comparison. All images of the selected classes in the training set and validation set are used to train and test the models, respectively.

\textbf{Baseline Methods and Models:}
\textcolor{crimson}{We compare our framework against seven state-of-the-art methods, which are Soft Filter Pruning (SFP) \cite{SoftPrune}, Filter Pruning via Geometric Median (FPGM) \cite{FPGM}, Network Slimming (NS) \cite{NetworkSlimming}, Prior Gradient Mask Guided Pruning-Aware Fine-Tuning (PGMPF) \cite{PGMPF}, Only Train Once (OTO) \cite{oto}, EfficientNet Compound  Scaling (Eff-Compd) \cite{tan2019efficientnet} and HACScale \cite{kong2022hacscale},} on four representative DNN architectures: VGGNet \cite{vgg}, Res-Net \cite{resnet}, DenseNet \cite{densenet}, and Goo-gLeNet \cite{googlenet}. The reason for using these networks is that their architectures use different kinds of network modules, which are popular and widely used. VGGNet has a simple sequential structure. ResNet, DenseNet, and GoogLeNet are the representatives of using Residual modules, Dense modules, and Inception modules, respectively. The architectures of other models are composed by these modules or their variants. On the CIFAR-10 dataset, we choose VGG-19, ResNet-164, and Dens-eNet-40 networks, and on the ImageNet-100 data-set, we choose VGG-19, GoogLeNet, and ResNet-50\footnote{In this paper, we keep the order of BN layers and convolution layers in the Residual modules.} networks. 

\textbf{Implementation Details:}
The proposed framework is implemented using PyTorch framework \cite{pytorch}. We set the start epoch automatically and set the Zero-Recovery interval to $2$. The training scheme for the CIFAR-10 dataset follows \cite{NetworkSlimming}, where the initial learning rate is $0.1$ with a decrease by the factor of $10$ at $50\%$ and $75\%$ of the total epochs. And the batch size is set to 64. For the ImageNet-100 dataset, following \cite{pytorch}, the batch size is 128, the initial learning rate starts at $0.1$ for ResNet-50 and GoogLeNet, and 0.01 for VGG-19, and the learning rate decays by a factor of $10$ every $1/3$ of all epochs. And for both datasets, the optimizer is SGD with a momentum of $0.9$ and weight decay of $1e^{-4}$. These settings are also used in Fig. \ref{fig:ks}. \textcolor{crimson}{ Due to comparison methods -- SFP \cite{SoftPrune}, FPGM \cite{FPGM} and PGMPF \cite{PGMPF} train models for 160 epochs on the CIFAR-10 dataset and 90 epochs on the ImageNet-100 dataset, respectively, we also choose these two settings of epochs for CIFAR-10 and ImageNet-100. And the comparison method -- NS \cite{NetworkSlimming} trains models for 320 epochs on the CIFAR-10 dataset and 180 epochs on the ImageNet-100 dataset totally, we choose these settings for comparison to NS \cite{NetworkSlimming}. For OTO \cite{oto}, the model is gradually compressed during training, which means that when the pre-defined maximum compression ratio is not reached, the ratio increases as the training epoch increases, and when the pre-defined maximum compression ratio is reached, the ratio no longer increases. Thus, we first use OTO to train models for 160 epochs on the CIFAR-10 dataset and 90 epochs on the ImageNet-100  dataset. If it meets the pre-defined compression ratio, we call it OTO-160/OTO-90 and compare it to methods that used the same training epochs. If it fails to meet the pre-defined compression ratio, we double its training epochs (named OTO-320/OTO-180) and compare it to methods with the currently same training epochs. Finally,  if it still does not meet the pre-defined compression ratio with the doubled epochs, its result is discarded.} Moreover, we need to compare the accuracy results under a specific latency constraint, hence we use open-source codes of these comparison methods to evaluate their performance.

\begin{table}[!t]
\centering
\scriptsize
\caption{\textcolor{crimson}{ Training time of VGG-19 model on ImageNet-100 via NVIDIA Quadro GV100 GPU.}}
\label{tab:overheadTime}
   \setlength{\tabcolsep}{6pt} 
    \renewcommand{\arraystretch}{1.1} %

\begin{tabular}{ccccc}
\hline
{\color[HTML]{000000} \textbf{Method}}  & {\color[HTML]{000000} \textbf{Epochs}} & {\color[HTML]{000000} \textbf{Time/Epoch(s)}} & {\color[HTML]{000000} \textbf{Total Time (h)}} & {\color[HTML]{000000} \textbf{\begin{tabular}[c]{@{}c@{}}Overhead\\ (\%)\end{tabular}}} \\ \hline
{\color[HTML]{000000} Trad.Training \cite{pytorch}}    & {\color[HTML]{000000} 90}              & {\color[HTML]{000000} 807}                    & {\color[HTML]{000000} 20.175}                  & {\color[HTML]{000000} 0.00}                                                             \\
{\color[HTML]{000000} SFP \cite{SoftPrune}}              & {\color[HTML]{000000} 90}              & {\color[HTML]{000000} 868}                    & {\color[HTML]{000000} 21.700}                  & {\color[HTML]{000000} 7.56}                                                             \\
{\color[HTML]{000000} FPGM \cite{FPGM}}             & {\color[HTML]{000000} 90}              & {\color[HTML]{000000} 896}                    & {\color[HTML]{000000} 22.400}                  & {\color[HTML]{000000} 11.03}                                                            \\
{\color[HTML]{000000} PGMPF \cite{PGMPF}}   & {\color[HTML]{000000} \textbf{90}}     & {\color[HTML]{000000} \textbf{1933}}          & {\color[HTML]{000000} \textbf{48.325}}         & {\color[HTML]{000000} \textbf{139.53}}                                                  \\
{\color[HTML]{000000} OTO-90 \cite{oto}}           & {\color[HTML]{000000} 90}              & {\color[HTML]{000000} 1180}                   & {\color[HTML]{000000} 29.500}                  & {\color[HTML]{000000} 46.22}                                                            \\
\rowcolor[HTML]{DBDBDB} 
{\color[HTML]{000000}\textbf{ Our-90}}           & {\color[HTML]{000000} 90}              & {\color[HTML]{000000} 816}                    & {\color[HTML]{000000} 20.400}                  & {\color[HTML]{000000} 1.12}                                                             \\ \hline
{\color[HTML]{000000} NS \cite{NetworkSlimming}}               & {\color[HTML]{000000} 180}             & {\color[HTML]{000000} 789}                    & {\color[HTML]{000000} 39.450}                  & {\color[HTML]{000000} 95.54}                                                            \\
{\color[HTML]{000000} OTO-180 \cite{oto}} & {\color[HTML]{000000} \textbf{180}}    & {\color[HTML]{000000} \textbf{1180}}          & {\color[HTML]{000000} \textbf{59.000}}         & {\color[HTML]{000000} \textbf{192.44}}                                                  \\
\rowcolor[HTML]{DBDBDB} 
{\color[HTML]{000000}\textbf{ Our-180}}          & {\color[HTML]{000000} 180}             & {\color[HTML]{000000} 816}                    & {\color[HTML]{000000} 40.800}                  & {\color[HTML]{000000} 102.23}                                                           \\ \hline
\end{tabular}
\end{table}

\subsection{\textcolor{crimson}{Training Overhead Evaluation}}
\label{overhead}
\textcolor{crimson}{
To demonstrate the high efficiency of our method, we evaluate the overhead of our method and other model pruning methods \cite{NetworkSlimming, SoftPrune, FPGM, PGMPF, oto}. As indicated in Table \ref{tab:overheadTime}, we train the VGG-19 model on the NVIDIA\textsuperscript{\textregistered} Quadro\textsuperscript{\textregistered} GV100 GPU using different methods, with the dataset ImageNet-100 and batch size set to $128$. \textit{Trad. Training} refers to traditional training that does not use a compact algorithm. NS \cite{NetworkSlimming} first pre-trains a full VGG-19 model before pruning this model. Then, it trains the small model after pruning for better accuracy. Thus, the average time cost per epoch of NS is less than that of {Trad. Training}. According to Table \ref{tab:overheadTime}, the overhead of \textit{\zbabb-90} is only $1.12\%$. Thus, under the same training epochs, the overhead of our compact learning can be neglected. However, the training overhead of other methods is much larger than that of our method. For example, the training overhead of PGMPF \cite{PGMPF} is about $139.53\%$, despite having the same training epochs as Trad Training. Moreover, its training overhead is even higher than that of \textit{\zbabb-180}, whose training epochs are double that of Trad. Training. In the following experiments, we place methods according to the number of training epochs. However, we need to note that even though PGMPF \cite{PGMPF} achieves the highest accuracy among all methods with the same training epochs in some cases, its training time is about double that of other methods.}

 \begin{figure}[t]
    \centering
    \includegraphics[width=0.48\textwidth]{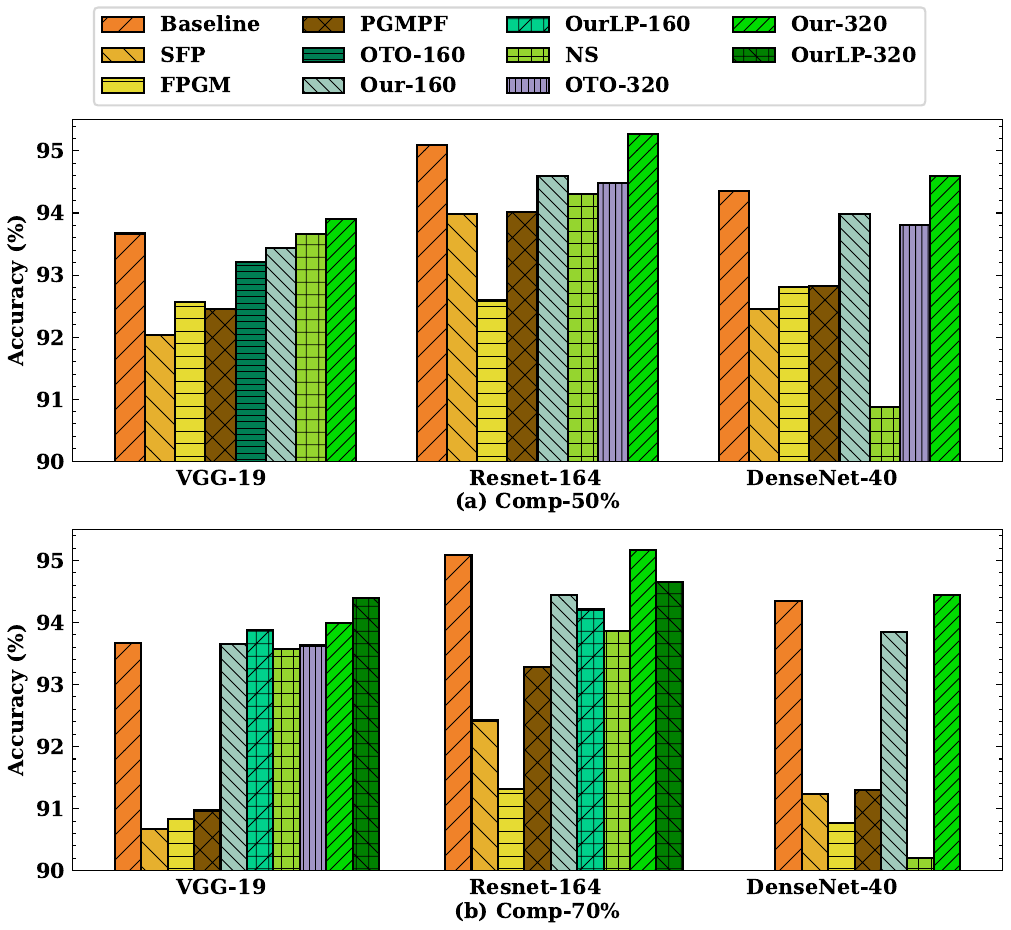}
    \caption{\textcolor{crimson}{Comparison of compact models on CIFAR-10.  Baseline means original models without compression. Omit \textit{{\zbabb}LP} if no layer pruning.}}
    \label{fig:cifar10}
\end{figure}
\subsection{Accuracy Evaluation}

We first evaluate our core compact learning in terms of accuracy on these datasets. This method can also compress a model by a specific compact ratio, only changing the $Comp\_ratio$ to a constant (\textit{Line 10} of Algorithm \ref{code:hdsp}).  Since the edge systems have a strict latency constraint, unlike other latency-agnostic approaches, larger compression ratios are required to verify the resilience of all methods. We employ a two aggressive compression ratio of 50\% (denoted as Comp-50\%) and 70\% (denoted as Comp-70\%) for evaluation.

\textbf{Results on the CIFAR-10 Dataset:} Fig. \ref{fig:cifar10} indicates that the our compression method achieves the highest accuracy (Acc.) \textcolor{crimson}{ compared to NS \cite{NetworkSlimming}, SFP \cite{SoftPrune}, FPGM \cite{FPGM}, PGMPF \cite{PGMPF} and OTO \cite{oto} under different compression ratios for each model.  In comparison to SFP \cite{SoftPrune}, FPGM \cite{FPGM},  PGMPF \cite{PGMPF} and OTO-160 \cite{oto}, } all algorithms train for 160 epochs without fine-tuning, and our method, marked as \textit{{\zbabb}-160}, can minimize the loss of accuracy for both 50\% and 70\% compression ratio. Compared to NS \cite{NetworkSlimming}, which needs  160 epochs of pre-training and 160 epochs of fine-tuning, \textcolor{crimson}{and OTO-320 \cite{oto},} \textit{{\zbabb}-160} can obtain a compact model with half the time cost and maintain a smaller loss of accuracy under both compression ratios. In addition, if our method trains for the same number of epochs as NS \cite{NetworkSlimming} \textcolor{crimson}{and OTO-320 \cite{oto} do}, denoted as \textit{{\zbabb}-320}, it can achieve higher accuracy than that of the original model, even at a compression ratio of 70\%, which shows the superiority of our framework. For layer pruning, although, in general, the accuracy of layer pruning is lower than that of channel pruning under the same compression ratio \cite{mao2017exploring}, our method, denoted as \textit{{\zbabb}LP}, can achieve similar accuracy as \textit{{\zbabb}} methods. Also, compared to other channel pruning methods \cite{NetworkSlimming, SoftPrune, FPGM, PGMPF}, its accuracy is higher than theirs.

\begin{table}[t]
\centering
\scriptsize
\caption{Comparison of compact models on ImageNet-100. Omit \textit{{\zbabb}LP} if no layer pruning.}
\label{tab:imagenet100}
   \setlength{\tabcolsep}{1.5pt} 
    \renewcommand{\arraystretch}{1.1} 
\begin{tabular}{cccccccccc}
\hline
                                 &                                            & \multicolumn{4}{c}{\textbf{Comp-50\% Acc. (\%)}}                                                                                                                                                                & \multicolumn{4}{c}{\textbf{Comp-70\% Acc. (\%)}}                                                                                                                                                                \\ \cline{3-10} 
\multirow{-2}{*}{\textbf{Model}} & \multirow{-2}{*}{\textbf{Method}}          & \textbf{Top-1}                         & \textbf{\begin{tabular}[c]{@{}c@{}}Top-1\\ Drop\end{tabular}} & \textbf{Top-5}                         & \textbf{\begin{tabular}[c]{@{}c@{}}Top-5\\ Drop\end{tabular}} & \textbf{Top-1}                         & \textbf{\begin{tabular}[c]{@{}c@{}}Top-1\\ Drop\end{tabular}} & \textbf{Top-5}                         & \textbf{\begin{tabular}[c]{@{}c@{}}Top-5\\ Drop\end{tabular}} \\ \hline
                                 & SFP  \cite{SoftPrune}                                        & 82.82                                  & 2.42                                                          & 95.34                                  & 0.82                                                          & 79.14                                  & 6.10                                                          & 93.50                                  & 2.66                                                          \\
                                 & FPGM  \cite{FPGM}                                        & 83.46                                  & 1.78                                                          & 95.48                                  & 0.68                                                          & 76.40                                  & 8.84                                                          & 92.10                                  & 4.06                                                          \\
                                & {\color[HTML]{000000} PGMPF \cite{PGMPF}}  & {\color[HTML]{000000}83.60} & {\color[HTML]{000000}1.64} & {\color[HTML]{000000}95.14} & {\color[HTML]{000000}1.02 }& {\color[HTML]{000000}80.58} & {\color[HTML]{000000}4.66} & {\color[HTML]{000000}94.10} & {\color[HTML]{000000}2.06} \\ & \cellcolor[HTML]{DBDBDB}\textbf{{\zbabb}-90}    & \cellcolor[HTML]{DBDBDB}\textbf{84.06} & \cellcolor[HTML]{DBDBDB}\textbf{1.18}                         & \cellcolor[HTML]{DBDBDB}\textbf{95.52} & \cellcolor[HTML]{DBDBDB}\textbf{0.64}                         & \cellcolor[HTML]{DBDBDB}\textbf{81.88} & \cellcolor[HTML]{DBDBDB}\textbf{3.36   }                               & \cellcolor[HTML]{DBDBDB}94.86          & \cellcolor[HTML]{DBDBDB}1.30                                  \\
                                 & \cellcolor[HTML]{DBDBDB}\textbf{{\zbabb}LP-90}  & \cellcolor[HTML]{DBDBDB}N.A.           & \cellcolor[HTML]{DBDBDB}N.A.                                  & \cellcolor[HTML]{DBDBDB}N.A.           & \cellcolor[HTML]{DBDBDB}N.A.                                  & \cellcolor[HTML]{DBDBDB}81.72          & \cellcolor[HTML]{DBDBDB}3.52                                  & \cellcolor[HTML]{DBDBDB}\textbf{94.92} & \cellcolor[HTML]{DBDBDB}\textbf{1.24}                         \\
                                 & NS \cite{NetworkSlimming}                                         & 82.66                                  & 2.58                                                          & 94.80                                  & 1.36                                                          & 78.88                                  & 6.36                                                          & 93.50                                  & 2.66                                                          \\ &{\color[HTML]{000000}OTO-180 \cite{oto}} & {\color[HTML]{000000}86.08} &{\color[HTML]{000000} -0.84}  &{\color[HTML]{000000}96.48} &{\color[HTML]{000000} -0.32} & {\color[HTML]{000000}N.A.}  & {\color[HTML]{000000}N.A.}  &{\color[HTML]{000000} N.A. } &{\color[HTML]{000000} N.A.} \\
\multirow{-8}{*}{VGG-19}         & \cellcolor[HTML]{DBDBDB}\textbf{{\zbabb}-180}   & \cellcolor[HTML]{DBDBDB}\textbf{87.18} & \cellcolor[HTML]{DBDBDB}\textbf{-1.94}                        & \cellcolor[HTML]{DBDBDB}\textbf{97.04} & \cellcolor[HTML]{DBDBDB}\textbf{-0.88}                        & \cellcolor[HTML]{DBDBDB}\textbf{84.14} & \cellcolor[HTML]{DBDBDB}\textbf{1.10}                         & \cellcolor[HTML]{DBDBDB}\textbf{95.90} & \cellcolor[HTML]{DBDBDB}\textbf{0.26}                         \\ \hline
                                 & SFP  \cite{SoftPrune}                                        & 82.84                                  & 1.76                                                          & 95.10                                  & 1.08                                                          & 82.38                                  & 2.22                                                          & 94.92                                  & 1.26                                                          \\
                                 & FPGM  \cite{FPGM}                                        & 83.54                        & 1.06                                                & 95.58                                  & 0.60                                                          & 81.96                                  & 2.64                                                          & 95.12                                  & 1.06                                                          \\
                                 &{\color[HTML]{000000} PGMPF \cite{PGMPF}} &{\color[HTML]{000000} \textbf{84.18}}
                                 &{\color[HTML]{000000}\textbf{ 0.42}}
                                 &{\color[HTML]{000000} 95.60}
                                 &{\color[HTML]{000000} 0.58}
                                 &{\color[HTML]{000000} {83.50}}
                                 &{\color[HTML]{000000} {1.10}}
                                 &{\color[HTML]{000000} \textbf{95.54}}
                                 &{\color[HTML]{000000} \textbf{0.64}} \\&
                                 \cellcolor[HTML]{DBDBDB}\textbf{{\zbabb}-90}    & \cellcolor[HTML]{DBDBDB}83.42          & \cellcolor[HTML]{DBDBDB}1.18                                  & \cellcolor[HTML]{DBDBDB}\textbf{95.70} & \cellcolor[HTML]{DBDBDB}\textbf{0.48}                         & \cellcolor[HTML]{DBDBDB}\textbf{83.58} & \cellcolor[HTML]{DBDBDB}\textbf{1.02}                         & \cellcolor[HTML]{DBDBDB}95.12          & \cellcolor[HTML]{DBDBDB}1.06                                  \\
                                 & \cellcolor[HTML]{DBDBDB}\textbf{{\zbabb}LP-90}  & \cellcolor[HTML]{DBDBDB}82.94          & \cellcolor[HTML]{DBDBDB}1.66                                  & \cellcolor[HTML]{DBDBDB}95.42          & \cellcolor[HTML]{DBDBDB}0.76                                  & \cellcolor[HTML]{DBDBDB}82.12          & \cellcolor[HTML]{DBDBDB}2.48                                  & \cellcolor[HTML]{DBDBDB}{95.26} & \cellcolor[HTML]{DBDBDB}{0.92}                         \\
                                 & NS \cite{NetworkSlimming}                                         & 82.56                                  & 2.04                                                          & 95.08                                  & 1.10                                                          & 80.18                                  & 4.42                                                          & 94.28                                  & 1.90                                                          \\
                                 & \cellcolor[HTML]{DBDBDB}\textbf{{\zbabb}-180}   & \cellcolor[HTML]{DBDBDB}85.74          & \cellcolor[HTML]{DBDBDB}-1.14                                 & \cellcolor[HTML]{DBDBDB}96.32          & \cellcolor[HTML]{DBDBDB}-0.14                                 & \cellcolor[HTML]{DBDBDB}84.48          & \cellcolor[HTML]{DBDBDB}0.12                                  & \cellcolor[HTML]{DBDBDB}95.66          & \cellcolor[HTML]{DBDBDB}0.52                                  \\
\multirow{-8}{*}{ResNet-50}      & \cellcolor[HTML]{DBDBDB}\textbf{{\zbabb}LP-180} & \cellcolor[HTML]{DBDBDB}\textbf{85.78} & \cellcolor[HTML]{DBDBDB}\textbf{-1.18}                        & \cellcolor[HTML]{DBDBDB}\textbf{96.40} & \cellcolor[HTML]{DBDBDB}\textbf{-0.22}                        & \cellcolor[HTML]{DBDBDB}\textbf{84.82} & \cellcolor[HTML]{DBDBDB}\textbf{-0.22}                        & \cellcolor[HTML]{DBDBDB}\textbf{95.86} & \cellcolor[HTML]{DBDBDB}\textbf{0.32}                         \\ \hline
                                 & SFP  \cite{SoftPrune}                                        & 81.82                                  & 3.50                                                          & 95.10                                  & 1.28                                                          & 77.30                                  & 8.02                                                          & 93.24                                  & 3.14                                                          \\
                                 & FPGM  \cite{FPGM}                                        & 82.72                                  & 2.60                                                          & \textbf{95.46}                         & \textbf{0.92}                                                 & 77.48                       & 7.84                                               & 92.94                                  & 3.44                                                          \\
                                 &{\color[HTML]{000000} PGMPF \cite{PGMPF}} &{\color[HTML]{000000} 82.84}
                                 &{\color[HTML]{000000} 2.48}
                                 &{\color[HTML]{000000} 95.40}
                                 &{\color[HTML]{000000} 0.98}
                                 &{\color[HTML]{000000} \textbf{78.94}}
                                 &{\color[HTML]{000000} \textbf{6.38}}
                                 &{\color[HTML]{000000} \textbf{94.24}}
                                 &{\color[HTML]{000000} \textbf{2.14}} \\&{\color[HTML]{000000} OTO-90 \cite{oto}} &{\color[HTML]{000000} 82.16}
                                 &{\color[HTML]{000000} 3.16}
                                 &{\color[HTML]{000000} 95.38}
                                 &{\color[HTML]{000000} 1.00}
                                 &{\color[HTML]{000000} {78.56}}
                                 &{\color[HTML]{000000} {6.76}}
                                 &{\color[HTML]{000000} {93.94}}
                                 &{\color[HTML]{000000} {2.44}} \\& \cellcolor[HTML]{DBDBDB}\textbf{{\zbabb}-90}    & \cellcolor[HTML]{DBDBDB}\textbf{82.86} & \cellcolor[HTML]{DBDBDB}\textbf{2.46 }                       & \cellcolor[HTML]{DBDBDB}95.44          & \cellcolor[HTML]{DBDBDB}0.94                                  & \cellcolor[HTML]{DBDBDB}77.28          & \cellcolor[HTML]{DBDBDB}8.04                                  & \cellcolor[HTML]{DBDBDB}{93.34} & \cellcolor[HTML]{DBDBDB}{3.04}                         \\
                                 & NS \cite{NetworkSlimming}                                         & 81.58                                  & 3.74                                                          & 95.02                                  & 1.36                                                          & N.A.                                   & N.A.                                                          & N.A.                                   & N.A.                                                          \\
\multirow{-7}{*}{GoogLeNet}      & \cellcolor[HTML]{DBDBDB}\textbf{{\zbabb}-180}   & \cellcolor[HTML]{DBDBDB}\textbf{84.48} & \cellcolor[HTML]{DBDBDB}\textbf{0.84}                         & \cellcolor[HTML]{DBDBDB}\textbf{95.92} & \cellcolor[HTML]{DBDBDB}\textbf{0.46}                         & \cellcolor[HTML]{DBDBDB}\textbf{79.20}  & \cellcolor[HTML]{DBDBDB}\textbf{6.12}                         & \cellcolor[HTML]{DBDBDB}\textbf{94.28} & \cellcolor[HTML]{DBDBDB}\textbf{2.10}                         \\ \hline
\end{tabular}

\end{table}

\textbf{Results on the ImageNet-100 Dataset:} Table \ref{tab:imagenet100} indicates that on a large-scale dataset, the proposed framework \textcolor{crimson}{ can also obtain the highest accuracy compared to NS \cite{NetworkSlimming}, SFP \cite{SoftPrune},  FPGM \cite{FPGM}, PGMPF \cite{PGMPF} and OTO \cite{oto} for most network architectures.} All lowest accuracy reduction (or highest accuracy increase) are marked in \textbf{boldface}.
\textcolor{crimson}{When applied to large models like VGG-19 and compared to SFP \cite{SoftPrune}, FPGM \cite{FPGM}, and PGMPF \cite{PGMPF}, all methods are trained for 90 epochs without fine-tuning, our method, denoted as \textit{{\zbabb}-90}, can achieve the highest accuracy under both 50\% and 70\% compression ratios. In the case of middle models such as ResNet-50, the average performance of \textit{ {\zbabb}-90} at 50\% and 70\% compression ratios is comparable to that of PGMPF \cite{PGMPF} and is better than SFP \cite{SoftPrune} and FPGM \cite{FPGM}. But the training time of \textit{ {\zbabb}-90} is much less than that of PGMPF \cite{PGMPF} (see Section \ref{overhead}). And for small models like GoogLeNet,  under the small compression ratio like 50\%,\textit{ {\zbabb}-90} can still minimize the loss of accuracy. But if 70\% of channels are pruned, the accuracy drop with all methods is large, as the redundancy of GoogLeNet is low. Compared to NS \cite{NetworkSlimming} with fine-tuning and OTO-180 \cite{oto}, when trained for the same number of epochs as theirs, our method, denoted as \textit{{\zbabb}-180}, can further minimize the loss of accuracy, and even outperform that of the original model, like VGG-19 and ResNet-50 under a compression ratio of 50\%. When considering the training time, \textit{{\zbabb}-180} can achieve higher accuracy than that of PGMPF \cite{PGMPF} with less time.}  Meanwhile, combined with layer pruning, our method, denoted as \textit{{\zbabb}LP}, can achieve similar accuracy as original \textit{{\zbabb}} methods, which means our layer pruning also works well for large datasets.


 \begin{figure}[t]
    \centering
    \includegraphics[width=0.48\textwidth]{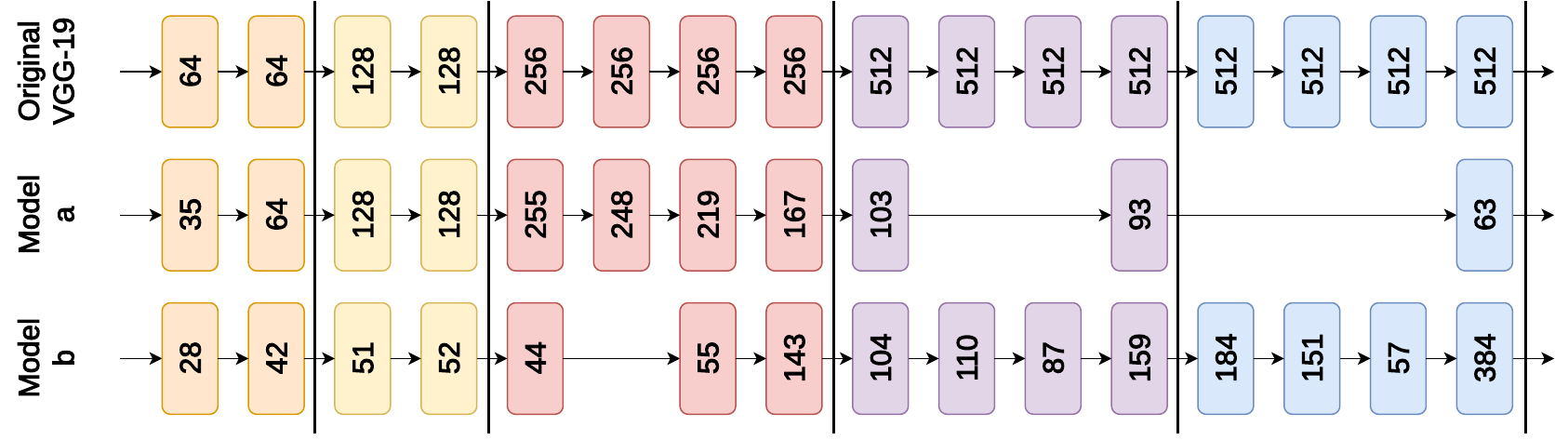}
    \caption{Model architecture from layer pruning of VGG-19. \textbf{Model a} is for Cifar-10 and \textbf{Model b} is for ImageNet-100. The vertical line represents the pooling layer, the rectangle represents the Conv-BN-ReLU layer, and the number in the rectangle represents the number of channels.}
    \label{fig:lparch}
\end{figure}

\textbf{Layer Pruning Architecture:} We visualize the layer pruned model architecture (eliminating 50\% channels) of VGG-19 by our method under different datasets in Fig. \ref{fig:lparch}, from which we observe that our engine encourages to train the desired architecture that goes deeper for a larger dataset.

\subsection{Latency-Accuracy Evaluation}
\label{latency-evaluation}
To demonstrate the effectiveness of our method for latency-critical edge systems, we evaluate the accuracy of our proposed framework and state-of-the-art methods under the $34ms$ latency constraint, which is equal to the recognition speed of human eyes -- $30$ frames per second (FPS). Table \ref{tab:imagenet100-latency} presents the results of proposed learning method compared with NS \cite{NetworkSlimming}, SFP \cite{SoftPrune}, FPGM \cite{FPGM}, \textcolor{crimson}{ PGMPF \cite{PGMPF}, OTO \cite{oto}, Eff-Compd \cite{tan2019efficientnet}, and HACScale \cite{kong2022hacscale}, }in which \textit{Q}-flag means with quantization, \textit{S}-flag means with model scaling and \textit{LP}-flag means with layer pruning. \textcolor{crimson}{The originally proposed compact learning method, denoted as \textit{{\zbabb}}, under the same training epoch as the original model, already achieves similar accuracy to the most recent works for all these three DNN models with much less training time (See Section \ref{overhead}). When equipped with layer pruning, denoted as  \textit{{\zbabb}LP},  its accuracy is better, and it can achieve the highest accuracy for most models. Layer pruning gives our framework an opportunity to find a more efficient architecture under the latency constraint. Particularly, when our method is trained for the same epoch as NS \cite{NetworkSlimming} and OTO-180 \cite{oto}, our method can exceed the accuracy of the uncompressed model a lot, whether equipped with layer pruning or not.} 

For the \textit{{\zbabb}Q} method, quantization can reduce the latency, hence we can use a smaller compression ratio in our compact learning. And after quantization, the compact model still meets the same latency constraint. It can achieve higher accuracy than the original \textit{{\zbabb}} method and obtain the best accuracy for all the three DNN architectures on different hardware platforms. 

For the \textit{{\zbabb}S} method, in these three DNN models, only the uncompressed GoogLeNet inference on NVIDIA Jetson TX2 consumes less time than the latency constraint, hence we uniformly scale up the model from width and depth, then compact it to meet the latency constraint by our compact learning. With the same training epochs to the unified scaling, Eff-Compd \cite{tan2019efficientnet}, \textcolor{crimson}{ and HACScale \cite{kong2022hacscale}, }we can achieve the highest accuracy. \textcolor{crimson}{ Layer pruning also provides our model scaling method to endue different implicit scaling factors (the number of non-zero $\gamma$ in Zero training) to depth and width, and these scaling factors are dynamically optimized during the learning process, this is the reason that our accuracy is higher than other methods.}


Additionally, each model in Table \ref{tab:imagenet100-latency} achieves a latency in the range of $33.2\pm 0.7 ms$, however, the compression ratio (negative ratios indicate scaling) of each model is not proportional to the reduction of latency. Meanwhile, the latency drop only has a weak relationship to the FLOPs drop. For example, the FLOPs drop of compact VGG-19 model under SFP \cite{SoftPrune} method is about $81.06\%$, which that of NS \cite{NetworkSlimming} method is about $76.59\%$, but they achieve the same latency. This is good proof of the necessity of the proposed framework to directly optimize latency for latency-critical edge systems. Furthermore, different methods have different attempt times to meet the latency constraint. For other methods, we used dichotomy to find the approximate compression/scaling ratio, but they still need about \textbf{$6-7$ times} to meet the specific latency constraint. However, with the assistance of the latency predictor, our method only needs one time. Therefore, a latency predictor is also critical in the DNN optimization for latency-critical edge systems.

\begin{table}[t]
\centering
\scriptsize
\caption{Comparison of optimized models on ImageNet-100 (s.t. 34ms). Omit \textit{{\zbabb}LP} if no layer pruning.}
\label{tab:imagenet100-latency}
   \setlength{\tabcolsep}{2.7pt} 
    \renewcommand{\arraystretch}{1.1} 
\begin{tabular}{cccccccc}
\hline
                                                                                 & \textbf{Model}                                                                 &                                            & \multicolumn{4}{c}{\textbf{Accuracy (\%)}}                                                                                                                                                                      &                                                                                       \\ \cline{2-2} \cline{4-7}
\multirow{-2}{*}{\textbf{Devices}}                                               & \textbf{\begin{tabular}[c]{@{}c@{}}Latency\\ (ms)\end{tabular}}                & \multirow{-2}{*}{\textbf{Method}}          & \textbf{Top-1}                         & \textbf{\begin{tabular}[c]{@{}c@{}}Top-1\\ Drop\end{tabular}} & \textbf{Top-5}                         & \textbf{\begin{tabular}[c]{@{}c@{}}Top-5\\ Drop\end{tabular}} & \multirow{-2}{*}{\textbf{\begin{tabular}[c]{@{}c@{}}Comp.\\ Ratio (\%)\end{tabular}}} \\ \hline
                                                                                 &                                                                                & SFP  \cite{SoftPrune}                              & 81.22                                  & 4.02                                                          & 94.94                                  & 1.22                                                          & 57.00                                                                                 \\
                                                                                 &                                                                                & FPGM  \cite{FPGM}                               & 82.17                                  & 3.07                                                          & 95.20                                  & 0.96                                                          & 57.00                                                                                 \\
                                                                                 &          
                                          
                                          &
                                          {\color[HTML]{000000} PGMPF \cite{PGMPF}}  & {\color[HTML]{000000}82.90} & {\color[HTML]{000000}2.33} & {\color[HTML]{000000}94.98} & {\color[HTML]{000000}1.18 }& {\color[HTML]{000000}55.00}\\ &
                                                                                 & \cellcolor[HTML]{DBDBDB}\textbf{{\zbabb}-90}    & \cellcolor[HTML]{DBDBDB}83.18          & \cellcolor[HTML]{DBDBDB}2.06                                  & \cellcolor[HTML]{DBDBDB}95.38          & \cellcolor[HTML]{DBDBDB}0.78                                  & \cellcolor[HTML]{DBDBDB}60.69                                                         \\
                                                                                 &                                                                                & \cellcolor[HTML]{DBDBDB}\textbf{{\zbabb}Q-90}   & \cellcolor[HTML]{DBDBDB}83.44          & \cellcolor[HTML]{DBDBDB}1.80                                  & \cellcolor[HTML]{DBDBDB}95.40          & \cellcolor[HTML]{DBDBDB}0.76                                  & \cellcolor[HTML]{DBDBDB}56.78                                                         \\
                                                                                 &                                                                                & \cellcolor[HTML]{DBDBDB}\textbf{{\zbabb}LP-90}  & \cellcolor[HTML]{DBDBDB}\textbf{85.74} & \cellcolor[HTML]{DBDBDB}\textbf{-0.50}                        & \cellcolor[HTML]{DBDBDB}\textbf{96.66} & \cellcolor[HTML]{DBDBDB}\textbf{-0.50}                        & \cellcolor[HTML]{DBDBDB}52.14                                                         \\
                                                                                 &                                                                                & NS \cite{NetworkSlimming}                                 & 80.14                                  & 5.10                                                          & 94.22                                  & 1.94                                                          & 56.00                                                                                 \\
                                                                                 &                                                                                & \cellcolor[HTML]{DBDBDB}\textbf{{\zbabb}-180}   & \cellcolor[HTML]{DBDBDB}85.78          & \cellcolor[HTML]{DBDBDB}-0.54                                 & \cellcolor[HTML]{DBDBDB}96.68          & \cellcolor[HTML]{DBDBDB}-0.52                                 & \cellcolor[HTML]{DBDBDB}60.90                                                         \\
                                                                                 &                                                                                & \cellcolor[HTML]{DBDBDB}\textbf{{\zbabb}Q-180}  & \cellcolor[HTML]{DBDBDB}86.64          & \cellcolor[HTML]{DBDBDB}-1.40                                 & \cellcolor[HTML]{DBDBDB}96.66          & \cellcolor[HTML]{DBDBDB}-0.50                                 & \cellcolor[HTML]{DBDBDB}55.98                                                         \\
                                                                                 & \multirow{-10}{*}{\begin{tabular}[c]{@{}c@{}}VGG-19\\ \\ 119.98\end{tabular}}   & \cellcolor[HTML]{DBDBDB}\textbf{{\zbabb}LP-180} & \cellcolor[HTML]{DBDBDB}\textbf{87.10} & \cellcolor[HTML]{DBDBDB}\textbf{-1.86}                        & \cellcolor[HTML]{DBDBDB}\textbf{96.74} & \cellcolor[HTML]{DBDBDB}\textbf{-0.58}                        & \cellcolor[HTML]{DBDBDB}53.16                                                         \\ \cline{2-8} 
                                                                                 &                                                                                & SFP  \cite{SoftPrune}                              & 83.60                                  & 1.00                                                          & 95.12                                  & 1.06                                                          & 42.00                                                                                 \\
                                                                                 &                                                                              & FPGM  \cite{FPGM}                               & 83.92                                  & 0.68                                                          & 95.96                                  & 0.22                                                          & 42.00                                                                                 \\
                                                                                 &        &
                                          {\color[HTML]{000000} PGMPF \cite{PGMPF}}  & {\color[HTML]{000000}84.04} & {\color[HTML]{000000}0.56} & {\color[HTML]{000000}95.28} & {\color[HTML]{000000}0.90}& {\color[HTML]{000000}40.00}\\ &                                                                                & \cellcolor[HTML]{DBDBDB}\textbf{{\zbabb}-90}    & \cellcolor[HTML]{DBDBDB}83.80          & \cellcolor[HTML]{DBDBDB}0.80                                  & \cellcolor[HTML]{DBDBDB}95.58          & \cellcolor[HTML]{DBDBDB}0.60                                  & \cellcolor[HTML]{DBDBDB}41.02                                                         \\
                                                                                 &                                                                                & \cellcolor[HTML]{DBDBDB}\textbf{{\zbabb}Q-90}   & \cellcolor[HTML]{DBDBDB}\textbf{84.18} & \cellcolor[HTML]{DBDBDB}\textbf{0.42}                         & \cellcolor[HTML]{DBDBDB}\textbf{96.00} & \cellcolor[HTML]{DBDBDB}\textbf{0.18}                         & \cellcolor[HTML]{DBDBDB}38.99                                                         \\
                                                                                 &                                                                                & \cellcolor[HTML]{DBDBDB}\textbf{{\zbabb}LP-90}  & \cellcolor[HTML]{DBDBDB}83.74          & \cellcolor[HTML]{DBDBDB}0.86                                  & \cellcolor[HTML]{DBDBDB}95.18          & \cellcolor[HTML]{DBDBDB}1.00                                  & \cellcolor[HTML]{DBDBDB}37.44                                                         \\
                                                                                 &                                                                                & NS \cite{NetworkSlimming}                                 & 80.08                                  & 4.52                                                          & 94.04                                  & 2.14                                                          & 45.00                                                                                 \\
                                                                                 &                   &
                                          {\color[HTML]{000000} OTO-180 \cite{oto}}  & {\color[HTML]{000000}84.70} & {\color[HTML]{000000}-0.10} & {\color[HTML]{000000}96.26} & {\color[HTML]{000000}-0.08 }& {\color[HTML]{000000}37.00}\\ &                                                                     & \cellcolor[HTML]{DBDBDB}\textbf{{\zbabb}-180}   & \cellcolor[HTML]{DBDBDB}86.82          & \cellcolor[HTML]{DBDBDB}-2.22                                 & \cellcolor[HTML]{DBDBDB}96.36          & \cellcolor[HTML]{DBDBDB}-0.18                                 & \cellcolor[HTML]{DBDBDB}40.89                                                         \\
                                                                                 & \multirow{-10}{*}{\begin{tabular}[c]{@{}c@{}}ResNet-50\\ \\ 49.06\end{tabular}} & \cellcolor[HTML]{DBDBDB}\textbf{{\zbabb}Q-180}  & \cellcolor[HTML]{DBDBDB}\textbf{87.04} & \cellcolor[HTML]{DBDBDB}\textbf{-2.44}                        & \cellcolor[HTML]{DBDBDB}\textbf{96.80} & \cellcolor[HTML]{DBDBDB}\textbf{-0.62}                        & \cellcolor[HTML]{DBDBDB}39.27                                                         \\ \cline{2-8} 
                                                                                 &                                                                                & Unified                                & 85.44                                  & -0.12                                                         & 96.42                                  & -0.04                                                         & -37.51                                                                                \\
                                                                                       &                                                                                & Eff-compd \cite{tan2019efficientnet}                                  & 85.52                                  & -0.20                                                         & 96.45                                  & -0.07                                                         & N.A.                                                                                  \\
                                                                                 &                   &
                                          {\color[HTML]{000000} HACScale \cite{kong2022hacscale}}  & {\color[HTML]{000000}85.62} & {\color[HTML]{000000}-0.30} & {\color[HTML]{000000}96.43} & {\color[HTML]{000000}-0.05 }& {\color[HTML]{000000}N.A.}\\ &                                                                           & \cellcolor[HTML]{DBDBDB}\textbf{{\zbabb}SLP-90}   & \cellcolor[HTML]{DBDBDB}85.64          & \cellcolor[HTML]{DBDBDB}-0.32                                 & \cellcolor[HTML]{DBDBDB}96.48          & \cellcolor[HTML]{DBDBDB}-0.10                                 & \cellcolor[HTML]{DBDBDB}-23.76                                                        \\
                                                                                 &                                                                                & \cellcolor[HTML]{DBDBDB}\textbf{{\zbabb}SQLP-90}  & \cellcolor[HTML]{DBDBDB}\textbf{86.10} & \cellcolor[HTML]{DBDBDB}\textbf{-0.78}                        & \cellcolor[HTML]{DBDBDB}\textbf{96.48} & \cellcolor[HTML]{DBDBDB}\textbf{-0.10}                        & \cellcolor[HTML]{DBDBDB}-33.76                                                        \\
                                                                                 &                                                                                & \cellcolor[HTML]{DBDBDB}\textbf{{\zbabb}SLP-180}  & \cellcolor[HTML]{DBDBDB}87.46          & \cellcolor[HTML]{DBDBDB}-2.14                                 & \cellcolor[HTML]{DBDBDB}97.02          & \cellcolor[HTML]{DBDBDB}-0.64                                 & \cellcolor[HTML]{DBDBDB}-24.28                                                        \\
\multirow{-27}{*}{\begin{tabular}[c]{@{}c@{}}NVIDIA\\ Jetson\\ TX2\end{tabular}} & \multirow{-7}{*}{\begin{tabular}[c]{@{}c@{}}GoogLeNet\\ \\ 20.27\end{tabular}} & \cellcolor[HTML]{DBDBDB}\textbf{{\zbabb}SQLP-180} & \cellcolor[HTML]{DBDBDB}\textbf{88.10} & \cellcolor[HTML]{DBDBDB}\textbf{-2.78}                        & \cellcolor[HTML]{DBDBDB}\textbf{97.12} & \cellcolor[HTML]{DBDBDB}\textbf{-0.74}                        & \cellcolor[HTML]{DBDBDB}-35.30                                                        \\ \hline
                                                                                 &                                                                                & SFP  \cite{SoftPrune}                              & 84.24                                  & 1.08                                                          & 95.80                                  & 0.58                                                          & 16.00                                                                                 \\
                                                                                 &                                                                                & FPGM  \cite{FPGM}                               & 84.96                                  & 0.36                                                          & 95.92                                  & 0.46                                                          & 16.00                                                                                 \\
                                                                                 &                    &
                                          {\color[HTML]{000000} PGMPF  \cite{PGMPF}}  & {\color[HTML]{000000}85.20} & {\color[HTML]{000000}0.12} & {\color[HTML]{000000}96.22} & {\color[HTML]{000000}0.16 }& {\color[HTML]{000000}17.00}\\ &                       &
                                          {\color[HTML]{000000} OTO-90 \cite{oto}}  & {\color[HTML]{000000}84.36} & {\color[HTML]{000000}0.96} & {\color[HTML]{000000}95.82} & {\color[HTML]{000000}0.56 }& {\color[HTML]{000000}18.00}\\ &                                                                           & \cellcolor[HTML]{DBDBDB}\textbf{{\zbabb}-90}    & \cellcolor[HTML]{DBDBDB}85.18          & \cellcolor[HTML]{DBDBDB}0.14                                  & \cellcolor[HTML]{DBDBDB}\textbf{96.66} & \cellcolor[HTML]{DBDBDB}\textbf{-0.28}                        & \cellcolor[HTML]{DBDBDB}17.08                                                         \\
                                                                                 &                                                                                & \cellcolor[HTML]{DBDBDB}\textbf{{\zbabb}Q-90}   & \cellcolor[HTML]{DBDBDB}\textbf{85.28} & \cellcolor[HTML]{DBDBDB}\textbf{0.04}                         & \cellcolor[HTML]{DBDBDB}{96.44} & \cellcolor[HTML]{DBDBDB}-0.06                                 & \cellcolor[HTML]{DBDBDB}15.86                                                         \\
                                                                                 &                                                                                & NS \cite{NetworkSlimming}                                 & 85.06                                  & 0.26                                                          & 95.90                                  & 0.48                                                          & 17.00                                                                                 \\
                                                                                 &                                                                                & \cellcolor[HTML]{DBDBDB}\textbf{{\zbabb}-180}   & \cellcolor[HTML]{DBDBDB}86.86          & \cellcolor[HTML]{DBDBDB}-1.54                                 & \cellcolor[HTML]{DBDBDB}96.94          & \cellcolor[HTML]{DBDBDB}-0.56                                 & \cellcolor[HTML]{DBDBDB}17.10                                                         \\
\multirow{-9}{*}{\begin{tabular}[c]{@{}c@{}}NVIDIA\\ Jetson\\ Nano\end{tabular}} & \multirow{-9}{*}{\begin{tabular}[c]{@{}c@{}}GoogLeNet\\ \\ 40.32\end{tabular}} & \cellcolor[HTML]{DBDBDB}\textbf{{\zbabb}Q-180}  & \cellcolor[HTML]{DBDBDB}\textbf{86.92} & \cellcolor[HTML]{DBDBDB}\textbf{-1.60}                        & \cellcolor[HTML]{DBDBDB}\textbf{96.88} & \cellcolor[HTML]{DBDBDB}\textbf{-0.50}                        & \cellcolor[HTML]{DBDBDB}16.11                                                         \\ \hline
\end{tabular}

\end{table}

\section{Conclusion}
\label{section:conclusion}
In this work, we have presented a neural network learning framework for latency-critical edge systems to directly optimize the latency of DNN on a specific edge device. This framework mainly includes three parts: a compact learning scheme, a latency predictor and a scaling scheme. From our experiments, the latency predictor is accurate enough with only a 6.12\% error. The proposed Zero-Recovery compact learning scheme can achieve less accuracy drop even under a large compression ratio, and it can increase the accuracy if it is trained with the same epochs as NS \cite{NetworkSlimming} on both datasets. Compared to the typical 3-stage pipeline pruning methods, our compact learning method can achieve higher accuracy with less time consumption. In addition, we evaluated our full framework under a latency constraint of $34$ ms, and the results show the efficiency and necessity of our method. Equipped with model scaling, our method can effectively adjust all DNN models to satisfy the system latency constraint. It solves the mismatch problem between state-of-the-art models and most edge devices. For example, we can compact VGG-19 model from $119.98$ms to $34$ms and we can also scale GoogLeNet model from $20.27$ms to $34$ms. Compared to other methods, our approach can obtain the highest accuracy with less training time. Meanwhile, for the conventional hardware-agnostic optimization methods, generally, they need to adjust the indirect parameters iteratively and evaluate on the real platform repeatedly to satisfy the latency constraint, which causes a large overhead. Our latency predictor greatly speeds up this process. Furthermore, our framework provides an optimization method for deploying machine learning on edge systems, it is not only suitable for latency but also suitable for energy and memory.

\section*{CRediT authorship contribution statement}
\textbf{Shuo Huai:} Conceptualization, Methodology, Software, Writing – original draft. \textbf{Di Liu:} Validation, Data curation, Writing - Review \& Editing. \textbf{Hao Kong:}  Data curation, Writing - Review \& Editing. \textbf{Ravi Subramaniam:} Writing - Review \& Editing, Resources. \textbf{Christian Makaya:} Writing - Review \& Editing, Resources. \textbf{Qian Lin:} Writing - Review \& Editing, Resources. \textbf{Weichen Liu:} Writing - Review \& Editing, Supervision, Project administration. 

\section*{Declaration of competing interest}
The authors declare the following financial interests/personal relationships which may be considered as potential competing interests: Weichen Liu reports financial support was provided by National Research Foundation (NRF) Singapore.

\section*{Acknowledgment}

This study is supported under the RIE2020 Industry Alignment Fund – Industry Collaboration Projects (IAF-ICP) Funding Initiative, as well as cash and in-kind contribution from the industry partner, HP Inc., through the HP-NTU Digital Manufacturing Corporate Lab (I1801E0028).




\bibliographystyle{elsarticle-num} 
\bibliography{reference}


\end{document}